\documentclass{article}

\usepackage{PRIMEarxiv}

\usepackage[utf8]{inputenc} 
\usepackage[T1]{fontenc}    
\usepackage{hyperref}       
\usepackage{url}            
\usepackage{booktabs}       
\usepackage{amsfonts}       
\usepackage{nicefrac}       
\usepackage{microtype}      
\usepackage{lipsum}
\usepackage{fancyhdr}       
\usepackage{graphicx}       
\usepackage{enumitem}
\usepackage{caption}
\usepackage{wrapfig}
\usepackage{pdfpages}
\usepackage{amsmath}
\usepackage[affil-it]{authblk}  
\usepackage{hyperref} 
\usepackage{titlesec}

\graphicspath{{media/}}     
\title{Automatic Quantification of Serial PET/CT Images for Pediatric Hodgkin Lymphoma Patients Using a Longitudinally-Aware Segmentation Network}
\author[1,2]{Xin Tie}
\author[1]{Muheon Shin}
\author[1]{Changhee Lee}
\author[1,3]{Scott B. Perlman}
\author[1]{Zachary Huemann}
\author[2]{Amy J. Weisman}
\author[4,5]{\\Sharon M. Castellino}
\author[6,7]{Kara M. Kelly}
\author[8]{Kathleen M. McCarten}
\author[9]{Adina L. Alazraki}
\author[10,11]{Junjie Hu}
\author[1,3]{\\Steve Y. Cho}
\author[1]{Tyler J. Bradshaw}

\affil[1]{\hspace{0.8em}Department of Radiology, University of Wisconsin, Madison, WI, USA}
\affil[2]{\hspace{0.8em}Department of Medical Physics, University of Wisconsin, Madison, WI, USA}
\affil[3]{\hspace{0.8em}University of Wisconsin Carbone Comprehensive Cancer Center, Madison, WI, USA }
\affil[4]{\hspace{0.8em}Department of Pediatrics, Emory University School of Medicine, Atlanta, GA, USA}
\affil[5]{\hspace{0.8em}Aflac Cancer and Blood Disorders Center, Children's Healthcare of Atlanta, Atlanta, GA, USA}
\affil[6]{\hspace{0.8em}Department of Pediatric Oncology, Roswell Park Comprehensive Cancer Center, Buffalo, NY, USA.}
\affil[7]{\hspace{0.8em}Department of Pediatrics, University at Buffalo Jacobs School of Medicine and Biomedical Sciences, Buffalo, NY, USA}
\affil[8]{\hspace{0.8em}Pediatric Radiology, Imaging and Radiation Oncology Core Rhode Island, Lincoln, RI, USA}
\affil[9]{\hspace{0.8em}Department of Radiology, Emory University School of Medicine and Children's Healthcare of Atlanta, Atlanta, GA, USA}
\affil[10]{\hspace{0.8em}Department of Biostatistics and Medical Informatics, University of Wisconsin, Madison, WI, USA}
\affil[11]{\hspace{0.8em}Department of Computer Science, School of Computer, University of Wisconsin, Madison, WI, USA}

\begin{document}
\maketitle
\thispagestyle{plain}  

\begin{abstract}
\textbf{Purpose}: Automatic quantification of longitudinal changes in PET scans for lymphoma patients has proven challenging, as residual disease in interim-therapy scans is often subtle and difficult to detect. Our goal was to develop a longitudinally-aware segmentation network (LAS-Net) that can quantify serial PET/CT images for pediatric Hodgkin lymphoma patients. \\
\textbf{Materials and Methods}: This retrospective study included baseline (PET1) and interim (PET2) PET/CT images from 297 patients enrolled in two Children’s Oncology Group clinical trials (AHOD1331 and AHOD0831). LAS-Net incorporates longitudinal cross-attention, allowing relevant features from PET1 to inform the analysis of PET2. Model performance was evaluated using Dice coefficients for PET1 and detection F1 scores for PET2. Additionally, we extracted and compared quantitative PET metrics, including metabolic tumor volume (MTV) and total lesion glycolysis (TLG) in PET1, as well as qPET and  $\Delta$SUVmax in PET2, against physician measurements. We quantified their agreement using Spearman’s $\rho$ correlations and employed bootstrap resampling for statistical analysis.  \\
\textbf{Results}: LAS-Net detected residual lymphoma in PET2 with an F1 score of 0.606 (precision/recall: 0.615/0.600), outperforming all comparator methods (P<0.01). For baseline segmentation, LAS-Net achieved a mean Dice score of 0.772. In PET quantification, LAS-Net’s measurements of qPET, $\Delta$SUVmax, MTV and TLG were strongly correlated with physician measurements, with Spearman’s $\rho$ of 0.78, 0.80, 0.93 and 0.96, respectively. The quantification performance remained high, with a slight decrease, in an external testing cohort.\\
\textbf{Conclusion}: LAS-Net demonstrated significant improvements in quantifying PET metrics across serial scans, highlighting the value of longitudinal awareness in evaluating multi-time-point imaging datasets.
\end{abstract}


\section{Introduction}
Among pediatric cancers, Hodgkin lymphoma (HL) is a highly curable malignancy (1), with 5-year survival exceeding 90$\%$ for patients receiving combination chemotherapy, radiation, or combined treatment (2). Despite this, pediatric patients face a significant risk of long-term side effects from therapeutic toxicities. Emerging evidence suggests that early responders to treatment may benefit from de-escalated therapies (3). Several clinical trials have used response assessment on interim Fluorodeoxyglucose (18F-FDG) PET scans after two cycles of chemotherapy for risk stratification (2,4). Currently, PET response is assessed using visual evaluation criteria, such as the Deauville score (DS) based on the lesion with the most intense uptake (5). Compared to the qualitative assessment, quantitative PET metrics have shown promise in guiding lymphoma treatment strategies (6,7). However, its use often relies on manual lesion segmentation, which is difficult and time-consuming, and has been limited to clinical trial settings. Deep learning (DL) algorithms have the potential to overcome this limitation and enable automatic PET analysis. 

There have been extensive studies using DL to segment lymphoma (8–11) and extract quantitative metrics (12–14) in PET scans. However, existing algorithms focus on quantifying baseline tumor burden, overlooking the important role of interim PET in response assessment. Compared to baseline PET, analyzing interim PET poses significant challenges, as tumor uptake is often subtle and difficult to differentiate from confounding physiologic or inflammatory FDG activity. Physicians typically rely on cross comparison with baseline PET to identify residual lymphoma, but methods for incorporating this information to interim PET analysis remain underexplored. 

In this study, we aimed to develop a longitudinally-aware segmentation network (LAS-Net) for automatic quantification of serial PET/CT images, facilitating PET-adaptive therapy for pediatric HL patients. Central to our design is a dual-branch architecture: one branch dedicated to segmenting lymphoma in baseline PET, while the other detects residual lymphoma in interim PET. The model was trained using PET/CT images from multiple centers as part of a phase 3 clinical trial. To assess the performance of our method, we evaluated its detection performance in interim PET and its segmentation performance in baseline PET. Furthermore, we extracted various quantitative PET metrics and quantified their agreement with physician measurements. We compared LAS-Net to other methods, including those with and without the integration of baseline PET information. Lastly, we performed external testing using data from another multi-center clinical trial of pediatric HL. 

\section{Materials and Methods}
\label{sec:headings}
\subsection{Patient Cohort}
This retrospective study included patients from two Children’s Oncology Group (COG) clinical trials: AHOD1331 (ClinicalTrials.gov number, NCT02166463) (2) and AHOD0831 (NCT01026220) (4). Both are phase 3 trials of pediatric patients aged 2-21 diagnosed with high-risk HL. The AHOD1331 trial assessed the utility of incorporating Brentuximab Vedotin with chemotherapy while the AHOD0831 trial evaluated the effects of combination chemotherapy together with radiation therapy. Baseline and interim FDG PET/CT images were gathered and transferred from IROC Rhode Island to our institution under data use agreements. Retrospective analysis was approved by institutional review board with no requirement of additional consent from patients. Of the 600 patients enrolled in the AHOD1331 trial, 200 with complete PET/CT datasets were randomly selected and used as our internal cohort. Among the 166 patients from the AHOD0831 trial, 97 had complete PET/CT datasets, and these were used for external testing. 

\subsection{Data Labeling}
For the AHOD1331 dataset, three experienced physicians with board certification in nuclear medicine (NM) provided lesion-level annotations for both baseline and interim PET using a semi-automated workflow (LesionID, MIM Software, Cleveland, Ohio), following a multi-reader adjudication process. One physician (M.S. with 5 years of experience) labelled all 200 cases while the other physicians (S.B.P. and S.Y.C., both with over 15 years of experience) each adjudicated 100 of the cases, refining the annotations by adding, deleting, or modifying contours as necessary, which were then reviewed and confirmed by the first reader. To assess residual tumors on interim PET scans, readers assigned each lesion a score on a 5-point scale, using the same visual evaluation criteria as the Deauville criteria (5). It is referred to as lesion-level Deauville score (LDS) in this study. Any residual disease scoring 3 to 5 had an associated contour. All segmented lesions were labeled according to physician confidence (non-equivocal or equivocal). An equivocal lesion is defined as a lesion that the physician is unsure whether its PET uptake corresponds to lymphoma or is due to physiological activity. Annotators were trained using a labeling guide (described in Appendix S1).

For the AHOD0831 dataset, PET images for each patient were annotated by one of two board-certified NM physicians (J.K. and I.L., both with 5 years of experience) on Mirada XD (Oxford, UK) software as part of a prior research study (12,15). Table \ref{table:table1} summarizes the characteristics of these two datasets.

\begin{table}[h!]
\vspace{-10pt}
\centering
\caption{Demographics and clinical characteristics of our internal and external cohorts.}
\includegraphics[width=0.9\textwidth]{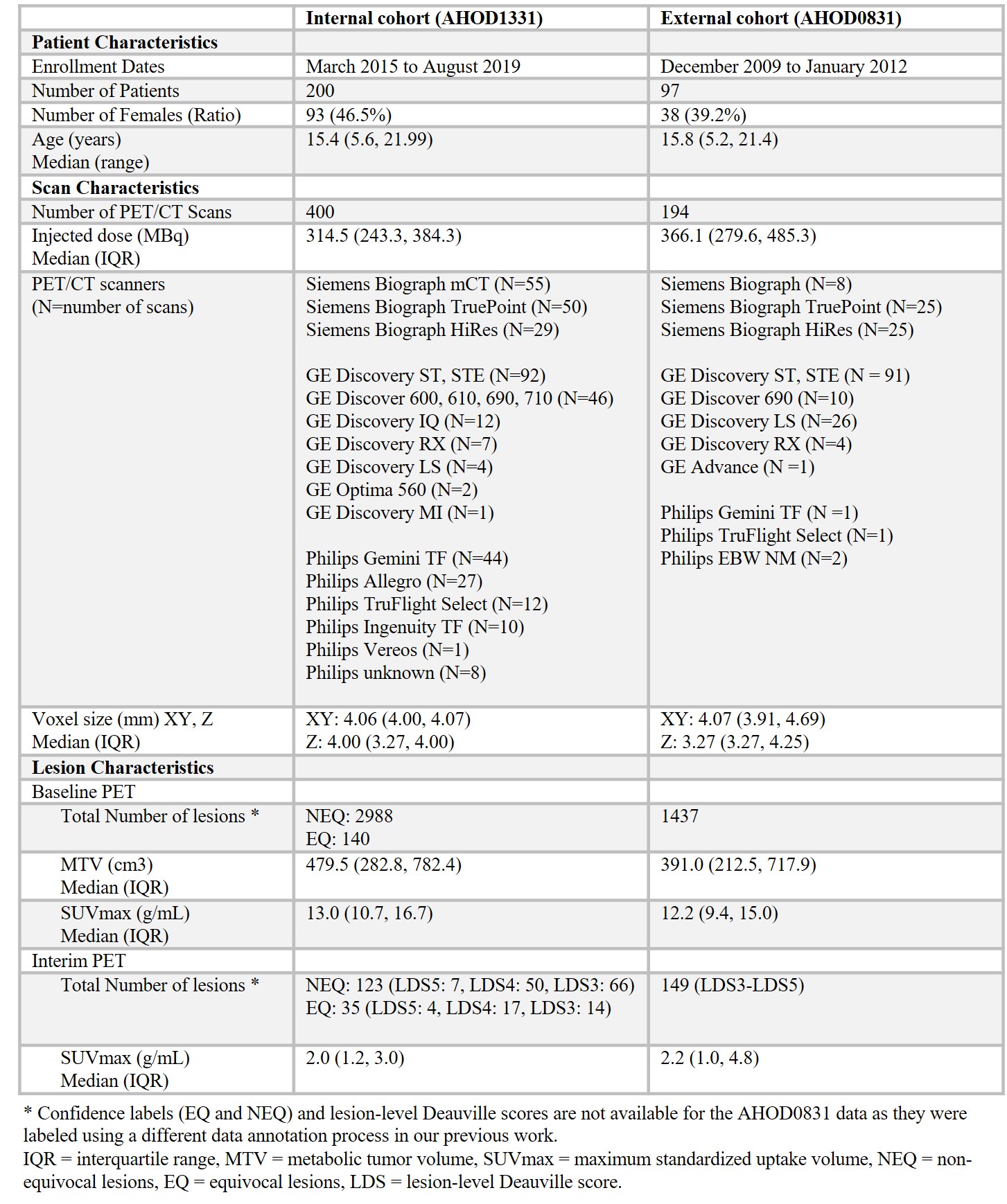}
\vspace{-5pt}
\label{table:table1}
\end{table}

\subsection{LAS-Net Architecture}
\begin{figure}[h!]
\centering
\includegraphics[width=0.995\textwidth]{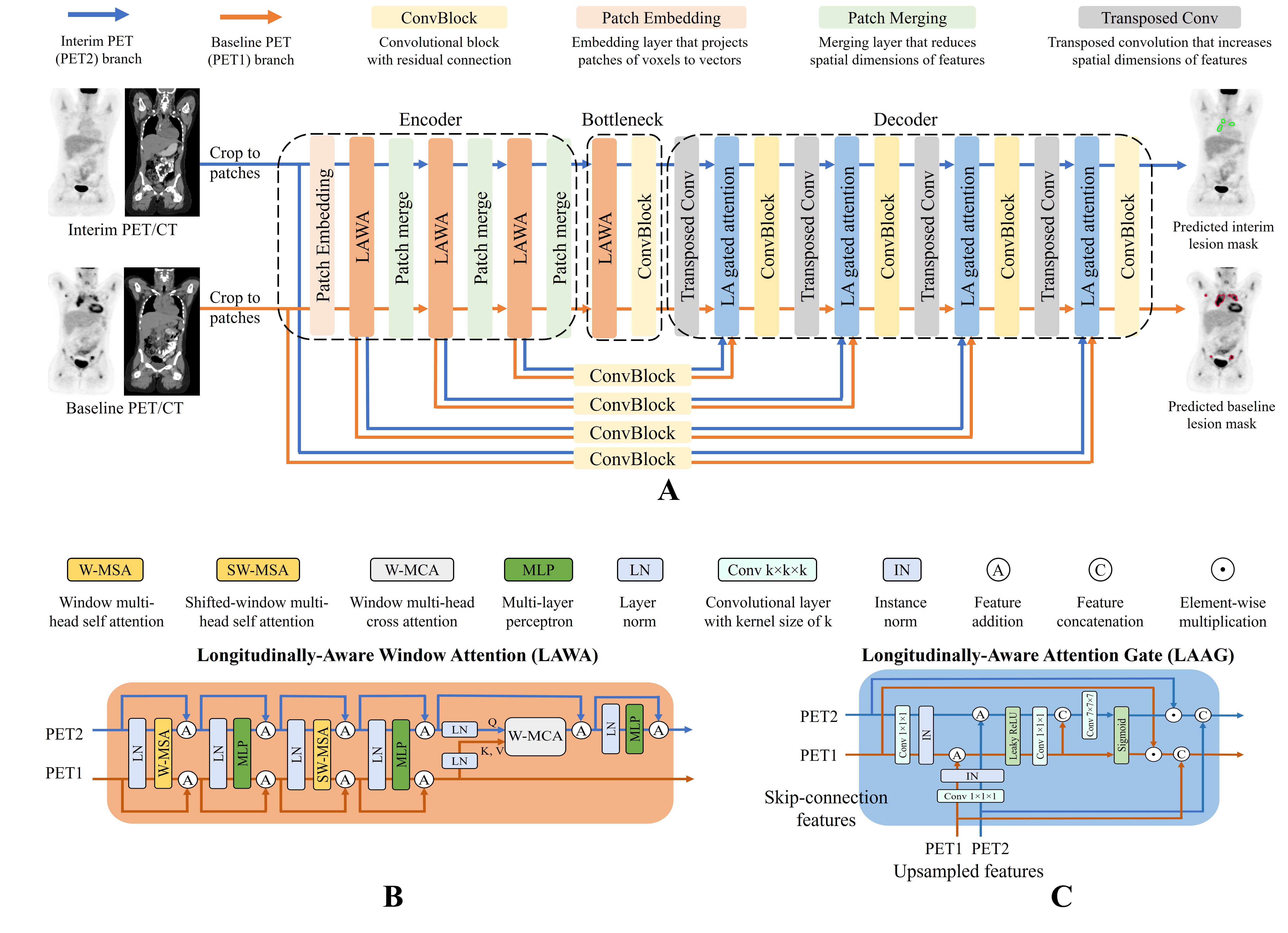}
  \caption{\small{The architecture of longitudinally-aware segmentation network (LAS-Net). (\textbf{A}) The dual-branch design accommodates baseline (PET1) and interim (PET2) PET/CT images. One branch is dedicated to processing PET1 while the other branch focuses on PET2, using features extracted from PET2 as well as the features from the PET1 branch. (\textbf{B}) The longitudinally-aware window attention (LAWA) module introduces multi-head cross-attention following two self-attention blocks. All attention layers have a window size of 7. (\textbf{C}) The longitudinally-aware attention gate (LAAG) introduces a learnable convolutional layer (kernel size=7) following the standard self-attention gate to refine the attention coefficients for PET2. Both LAWA and LAAG modules only allow one-way information flow from the PET1 to the PET2 branch.}} 
  \vspace{-5pt}
  \label{fig:fig1}
\end{figure}

We designed LAS-Net with a dual-branch architecture to accommodate baseline and interim PET/CT images, as illustrated in Figure \ref{fig:fig1}A. One branch exclusively processes baseline PET (PET1) and predicts the corresponding lesion masks. The other branch focuses on interim PET (PET2), but also utilizes information extracted from the PET1 branch to generate masks of residual lymphoma. This architecture enables our model to gather useful information from PET1 to inform and improve the analysis of subsequent scans. Meanwhile, it ensures a one-way information flow, preventing PET2 information from influencing PET1 analysis. 

Like many segmentation networks, LAS-Net was adapted from a UNet-like architecture. It is based on 3D SwinUNETR (16), a state-of-the-art (SOTA) model comprising a Swin Transformer (17) encoder and a convolutional neural network (CNN) decoder. In LAS-Net, each convolutional block is a stack of two convolution units (3$\times$3$\times$3 convolution sub-layers, instance normalization, leaky ReLU) with a residual connection. Beyond these components, we have introduced two critical mechanisms to allow information from the PET1 branch to influence the PET2 branch. One is the longitudinally-aware window attention (LAWA) on the encoder side, and the other is the longitudinally-aware attention gate (LAAG) on the decoder side.

Figure \ref{fig:fig1}B illustrates the structure of the LAWA module. Compared to the standard Swin Transformer block (17), this module introduces a window-based multi-head cross-attention (W-MCA) layer with a window size of 7$\times$7$\times$7 in the PET2 branch. The W-MCA takes the \textit{query} vectors from PET2 features and the \textit{key} and \textit{value} vectors from PET1 features. It computes the attention matrix of the \textit{query} and \textit{key} using scaled dot product, allowing the model to dynamically allocate focus based on the relevance of regions across PET1 and PET2. The \textit{value} vectors are then reweighted by this attention matrix and added to input PET2 features. 

Figure \ref{fig:fig1}C presents the design of the LAAG module. Similar to the original attention gate (18), the LAAG module processes inputs from both the prior layer and skip connections, generating attention coefficients. To enable additional longitudinal awareness, we concatenate the attention coefficients derived from PET1 and PET2 and convolve them with a learnable 7$\times$7$\times$7 kernel to refine the PET2 attention coefficients. This CNN-based cross-attention gate allows the LAAG module to select PET2 features using information from the PET1 branch. 

LAS-Net operates on 112$\times$112$\times$112 patches from co-registered baseline and interim PET/CT images. Except for the longitudinal cross-attention components, all other weights in the model are shared between the PET1 and PET2 branches. The model was jointly optimized for PET1 and PET2 lesion segmentation using a compound loss, comprised of cross-entropy and Dice loss. Models were trained and evaluated through fivefold cross-validation (N=40 in each fold). Implementation details can be found in Appendices S2-3.

\subsection{Quantitative PET Metrics}
In baseline PET analysis, we evaluated model performance using the Dice coefficient, false positive volume (FPV), and false negative volume (FNV) per patient. The quantitative metrics computed for PET1 scans (definitions in Appendix S4) included metabolic tumor volume (MTV) (19), total lesion glycolysis (TLG) (20), maximum lesion standardized uptake value (SUVmax), maximum tumor dissemination (Dmax) (21), maximum distance between the lesion and the spleen (Dspleen) (22) and the number of lesions. Since interim PET analysis primarily involves SUVmax or SUVpeak measurements (23), accurate tumor segmentation is not needed. Consequently, for PET2 scans, we evaluated our model’s performance using detection F1 scores, precision, and recall. The evaluation criterion for lesion detection is defined as follows: a predicted lesion was considered a true positive if it overlapped with any true lesion identified by the physician. A true lesion without overlap was classified as a false negative. Additionally, we included a more stringent criterion which required the SUVmax measured for the predicted lesion to be matched with the true lesion, otherwise the predicted lesion was classified as a false positive and the true lesion was classified as a false negative. Lesions detected by the model that were considered as equivocal by the physicians were not counted as false positives or true positives in our evaluation. We also extracted quantitative PET2 metrics from model predictions, including SUVmax, percentage difference between baseline and interim SUVmax ($\Delta$SUVmax), qPET (23), and the number of residual lesions. Notably, $\Delta$SUVmax and qPET have been demonstrated to have predictive potential for patient prognosis (23–25). The agreement between automated PET metrics and physician measurements was quantified by Spearman’s $\rho$ correlations.

\subsection{Model Comparison}
We compared the performance of LAS-Net to other models trained on our dataset, including DynUNet (26,27), SegResNet (28) and SwinUNETR (16). No longitudinal cross-attention was incorporated into these models’ architectures. We also evaluated Clinical Knowledge-Driven Hybrid Transformer (CKD-Trans) (29) and Spatial-Temporal Transformer (ST-Trans) (30), both of which integrated information from PET1 into PET2 analysis using cross-attention. Notably, CKD-Trans and ST-Trans were initially developed for tumor segmentation in multiparametric MRI. Table \ref{table:table2} summarizes key differences among these models.

Furthermore, we implemented a previous technique (15,31) that used deformable registration between PET1 and PET2 scans to reduce false positives in PET2 lesion masks. Specifically, segmentation masks predicted for PET1 are propagated to PET2 using deformable registration, and then PET2 contours that do not overlap with PET1 contours are excluded. In our work, we refer to this technique as “mask propagation through deformable registration” (MPDR). Quantitative results were reported both with and without MPDR. Additionally, we conducted ablation studies to assess the effectiveness of individual components in LAS-Net. 

\begin{table}[h!]
\centering
\caption{Characteristics of the models evaluated in this study. }
\includegraphics[width=0.98\textwidth]{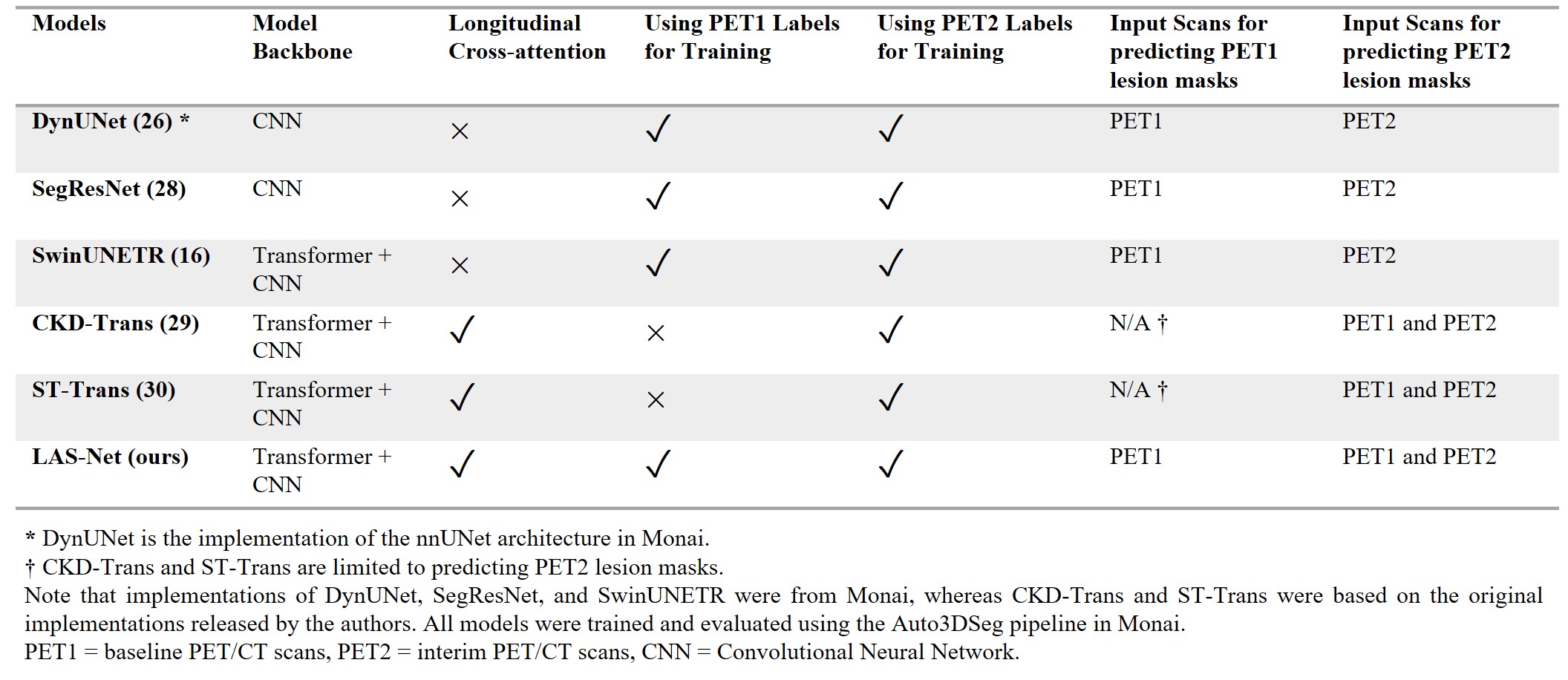}
  \vspace{-5pt}
\label{table:table2}
\end{table}

\subsection{Agreement of Predicted DSs and Physician-assigned DSs}
DSs serve as an internationally accepted scoring system for assessing treatment response in interim PET. Two types of thresholds (DS3-DS5 positive or DS4-DS5 positive) are often used to categorize patients into adequate or inadequate response classes, depending on the clinical context (e.g., certain trials focused on treatment de-escalation may consider DS3 as an inadequate response) (32). DS for each individual patient is determined by the residual lesion with the highest uptake (5). Although our model was not trained to output patient-level DSs, we can estimate DSs by converting extracted qPET values to DSs using the qPET criterion (23). In this context, the patient-level DS was inferred from the highest-valued LDS within the patient. This indirect method allowed for a comparison of model-predicted DSs and physician-assigned DSs. The level of agreement was quantified by the F1 score and the Kappa index.

\subsection{Statistical Analysis}
The 95$\%$ confidence intervals (CIs) for our results were derived using nonparametric bootstrap resampling (33) with 10,000 repetitive trials. Baseline and interim PET/CT scans were analyzed separately in all statistical evaluations. Specifically, bootstrap resampling was performed at the patient level, meaning that patients were sampled with replacement and each patient’s baseline and interim scans were included independently in their respective analyses. For each evaluation metric, the difference between LAS-Net and a comparator method was considered statistically significant at the 0.05 level if the metric values computed for LAS-Net exceeded those of the comparator method in 95$\%$ of trials.

\subsection{Data Availability}
The COG clinical trial data is archived in NCTN Data Archive. Our algorithm was implemented using the Auto3dSeg pipeline in Monai (27). The code and models are available in the open-source project: \url{https://github.com/xtie97/LAS-Net}.

\section{Results}
\subsection{Quantitative Performance}
\begin{figure}[ht!]
\centering
\includegraphics[width=0.98\textwidth]{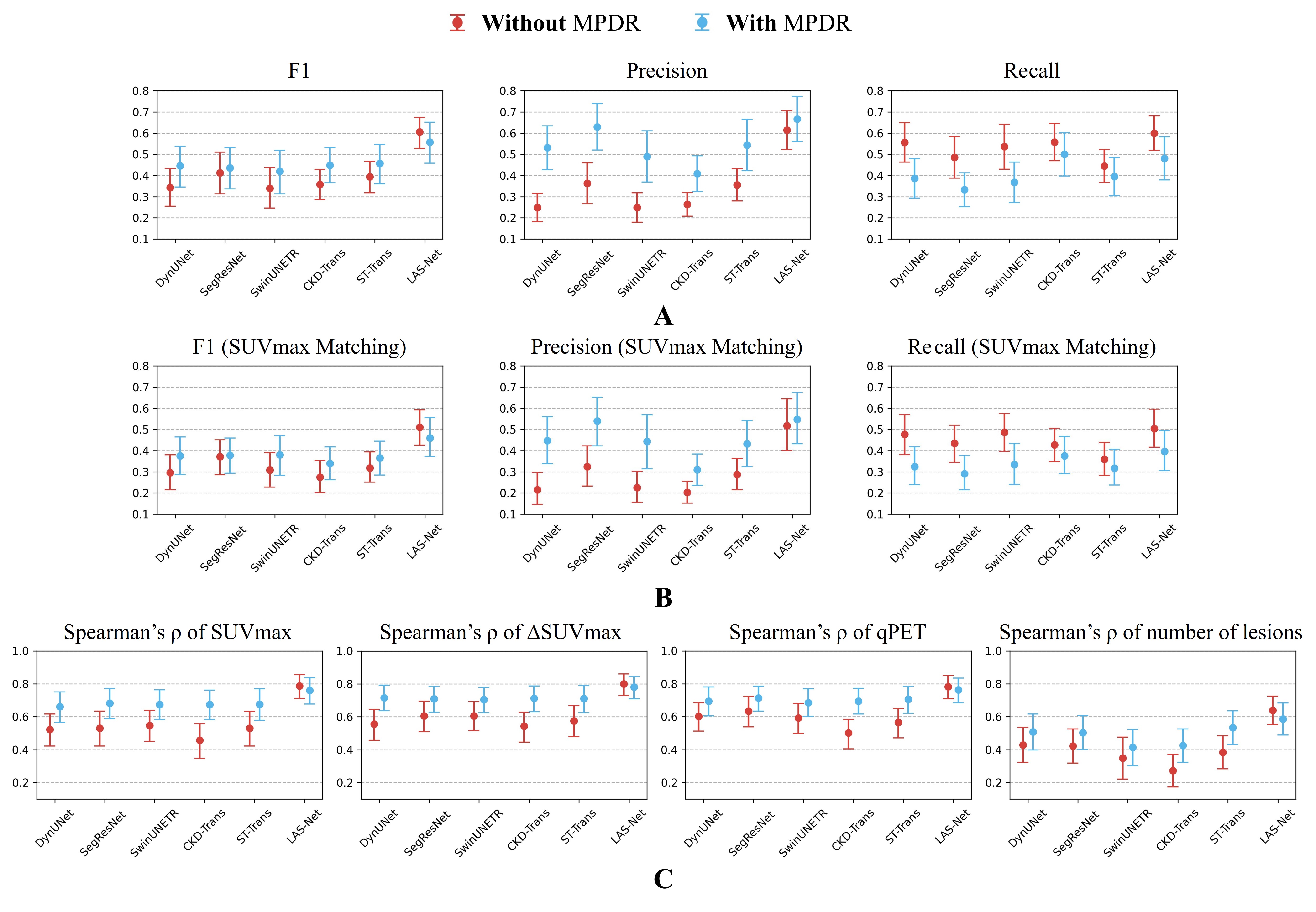}
  \caption{\small{Performance comparison of interim PET lesion detection in the internal cohort. Results are reported with and without mask propagation through deformable registration (MPDR). Notably, CKD-Trans and ST-Trans utilized baseline lesion masks predicted by DynUNet for MPDR. (\textbf{A}) and (\textbf{B}) present the results of detection F1 scores, precision, and recall using different criteria to classify true positives. In (\textbf{A}), a predicted lesion is classified as a true positive if it overlaps with at least one voxel of the reference lesion. In (\textbf{B}), a predicted lesion is considered a true positive if its SUVmax is matched with the reference lesion’s SUVmax. (\textbf{C}) quantifies the agreement between model predictions and physician measurements for interim PET metrics. In the plots, actual metric values and Spearman's correlation values are marked by circles with error bars indicating 95$\%$ confidence intervals. LAS-Net showed significantly improved performance ($P<0.05$) over all comparator methods in F1 scores and interim PET metrics. The only exception was for qPET where its performance did not significantly surpass SegResNet with MPDR ($P=0.057$). SUVmax = maximum lesion standardized uptake value, $\Delta$SUVmax = percentage difference of SUVmax between the baseline and interim scans.}} 
  \label{fig:fig2}
\end{figure}

Figure \ref{fig:fig2}A shows the comparison of lesion detection performance in PET2 across all evaluated models. LAS-Net detected residual lymphoma with an F1 score of 0.606 (95$\%$CI, 0.528, 0.674). Applying MPDR to predicted interim masks increased LAS-Net’s precision (0.615 to 0.667), but at the cost of reduced recall (0.600 to 0.481). This suggests that MPDR may filter out true positive lesions, including new lesions that are not present in the baseline scan. Overall, the use of MPDR did not improve the detection performance of LAS-Net (F1 without vs. with MPDR: 0.606 vs. 0.558, P=0.22). Conversely, all comparator models benefited from MPDR, with ST-Trans (with MPDR) achieving the highest F1 score (0.457, 95$\%$CI, 0.360, 0.547) of the comparator methods. Nevertheless, it was statistically inferior (P=0.005) to LAS-Net in identifying residual lesions. If requiring SUVmax to be matched (Figure \ref{fig:fig2}B), which is a stricter criterion, the F1 score of LAS-Net was 0.511 (95$\%$CI, 0.426, 0.592). With this criterion, it still surpassed all comparator methods. When equivocal lesions were included, LAS-Net’s detection F1 scores were 0.586 (95$\%$CI, 0.510, 0.660) for the overlapping criterion and 0.486 (95$\%$CI, 0.407, 0.560) for the SUVmax matching criterion (detailed comparison in Appendix S5). In terms of quantitative PET2 metrics, LAS-Net consistently outperformed other methods (Figure \ref{fig:fig2}C), with Spearman’s $\rho$ correlations of 0.79 (95$\%$CI, 0.70, 0.86) for SUVmax, 0.80 (95$\%$CI, 0.72 0.86) for $\Delta$SUVmax, 0.78 (95$\%$CI, 0.70, 0.85) for qPET and 0.64 (95$\%$CI, 0.54, 0.72) for the number of lesions. $\Delta$SUVmax and qPET had exact matches in values for 55$\%$ (110/200) and 69.5$\%$ (139/200) of cases, respectively.

\begin{figure}[h!]
\vspace{-5pt}
\centering
\includegraphics[width=0.8\textwidth]{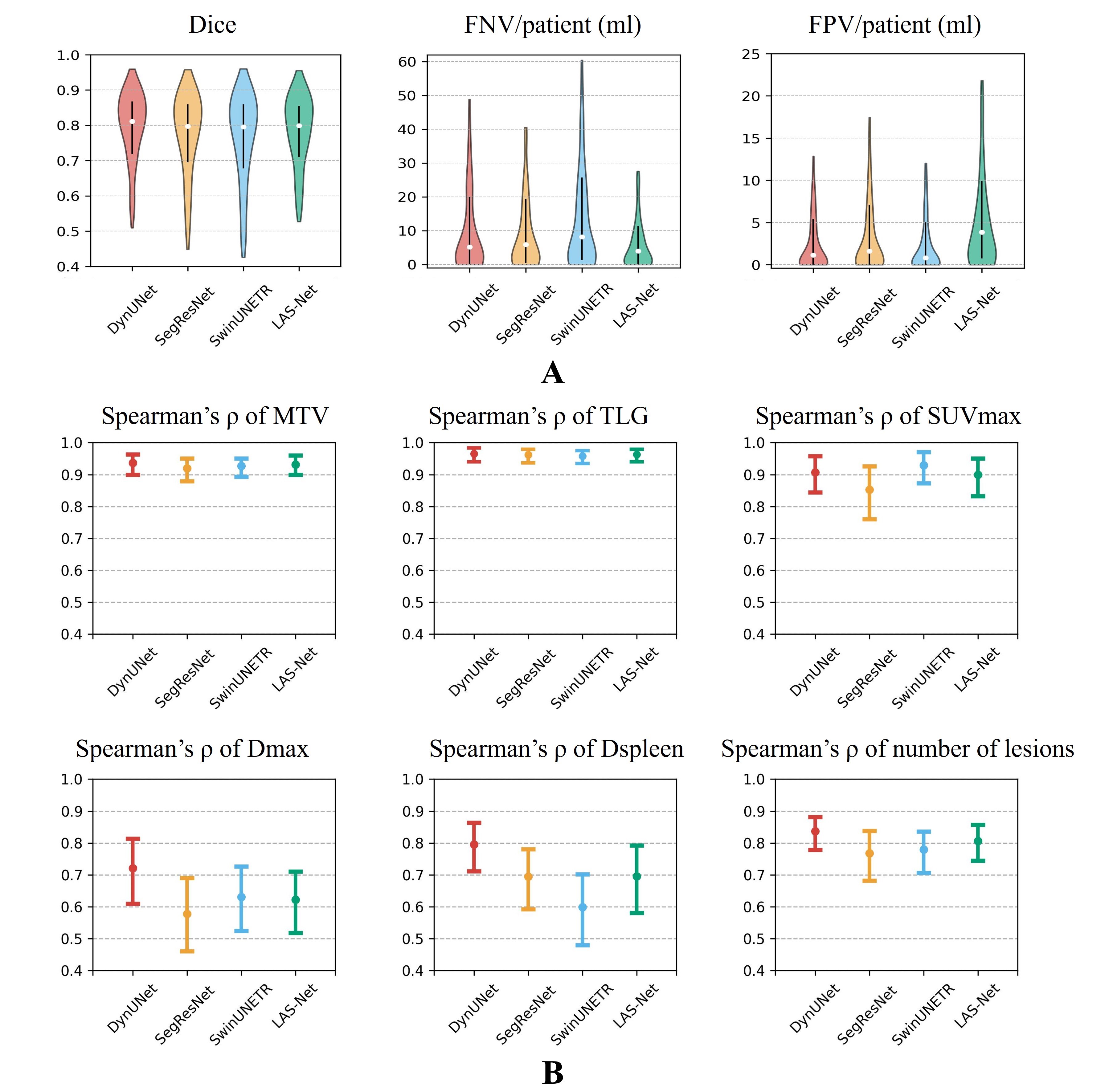}
  \caption{\small{Performance comparison of baseline PET lesion segmentation in the internal cohort. (\textbf{A}) shows violin plots of evaluation metrics, where vertical lines represent the interquartile ranges and white circles mark the median values. (\textbf{B}) compares the correlations between baseline PET metrics assessed by physicians and those measured by deep learning models. Actual Spearman’s correlation values are marked by circles and their 95$\%$ confidence intervals are denoted by error bars. FPV = false positive volume, FNV = false negative volume, MTV = metabolic tumor volume, TLG = total lesion glycolysis, SUVmax = maximum lesion standardized uptake value, Dmax = maximum tumor dissemination, Dspleen = maximum distance between the lesion and the spleen. }} 
    \vspace{-5pt}
  \label{fig:fig3}
\end{figure}

For automatic PET1 analysis (Figure \ref{fig:fig3}), LAS-Net attained a mean Dice score of 0.772 (95$\%$CI, 0.752, 0.791), with average FNV of 10.80 ml (95$\%$CI, 8.53, 13.46) and FPV of 9.68 ml (95$\%$CI, 7.50, 12.40) per patient. It demonstrated comparable performance to the best model, DynUNet, which had a Dice score of 0.779 (95$\%$CI, 0.758, 0.797, P=0.32). Among the PET1 metrics extracted by LAS-Net, MTV, TLG and SUVmax exhibited high correlations with the values measured by physicians ($\rho$=0.93 for MTV, 0.96 for TLG, 0.90 for SUVmax). No significant differences were observed across the four evaluated models for these metrics. For the distance-based metrics, Dmax and Dspleen, LAS-Net showed moderate correlations ($\rho$=0.62 for Dmax, 0.70 for Dspleen) with physician measurements, indicating the challenges of detecting individual lesions at the farthest distances. 

\begin{figure}[t!]
\vspace{-20pt}
\centering
\includegraphics[width=0.75\textwidth]{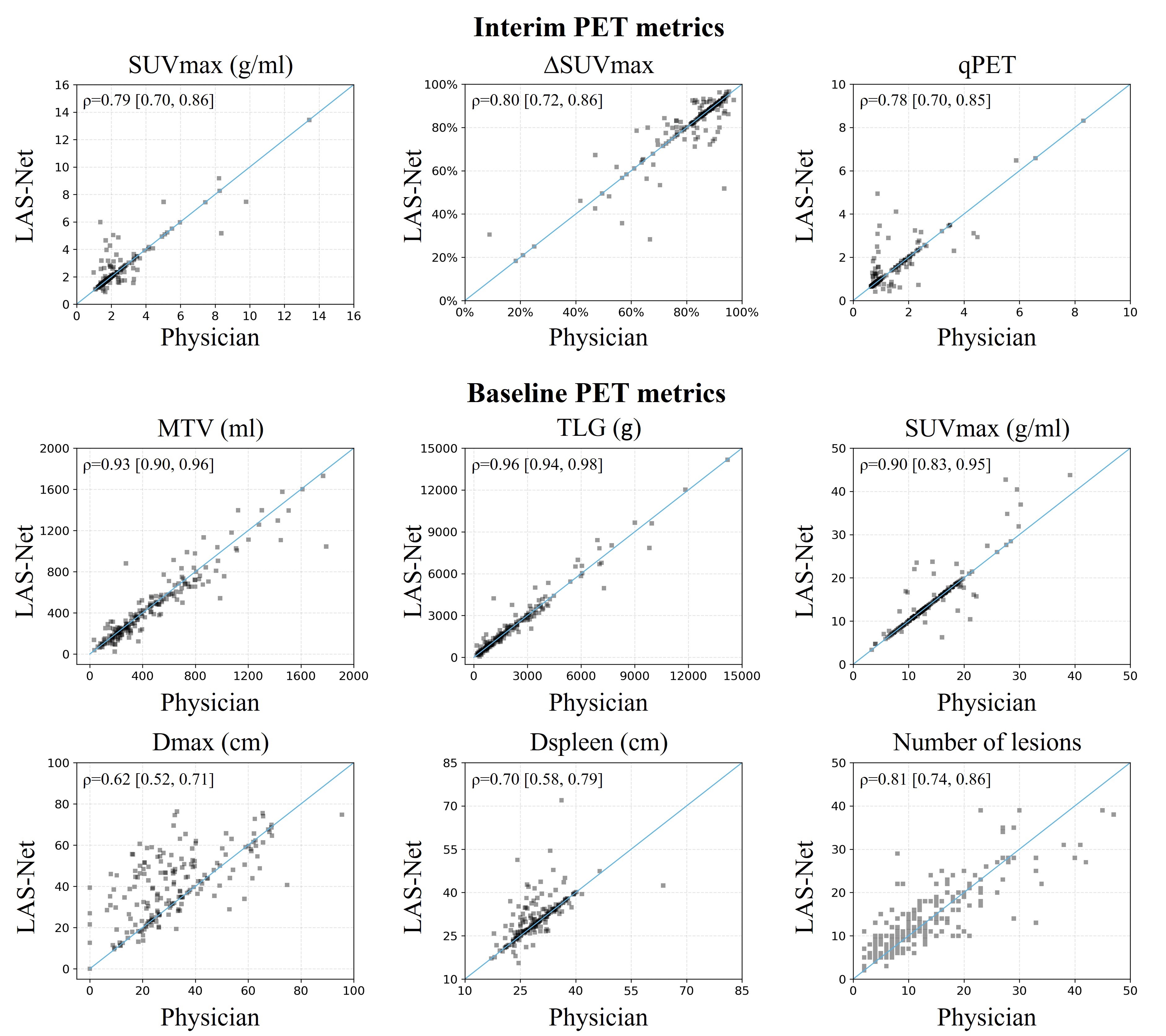}
  \caption{\small{Comparison of physician-based and automatically extracted PET metrics. Spearman’s $\rho$ correlations are shown in the top left corner of each plot. Correlation values are presented as mean [2.5th percentile, 97.5th percentile].}} 
  \label{fig:fig4}
\end{figure}

Scatter plots in Figure \ref{fig:fig4} visualize the agreement between PET metrics assessed by physicians and those measured by LAS-Net.

\begin{figure}[h!]
\centering
\includegraphics[width=0.9\textwidth]{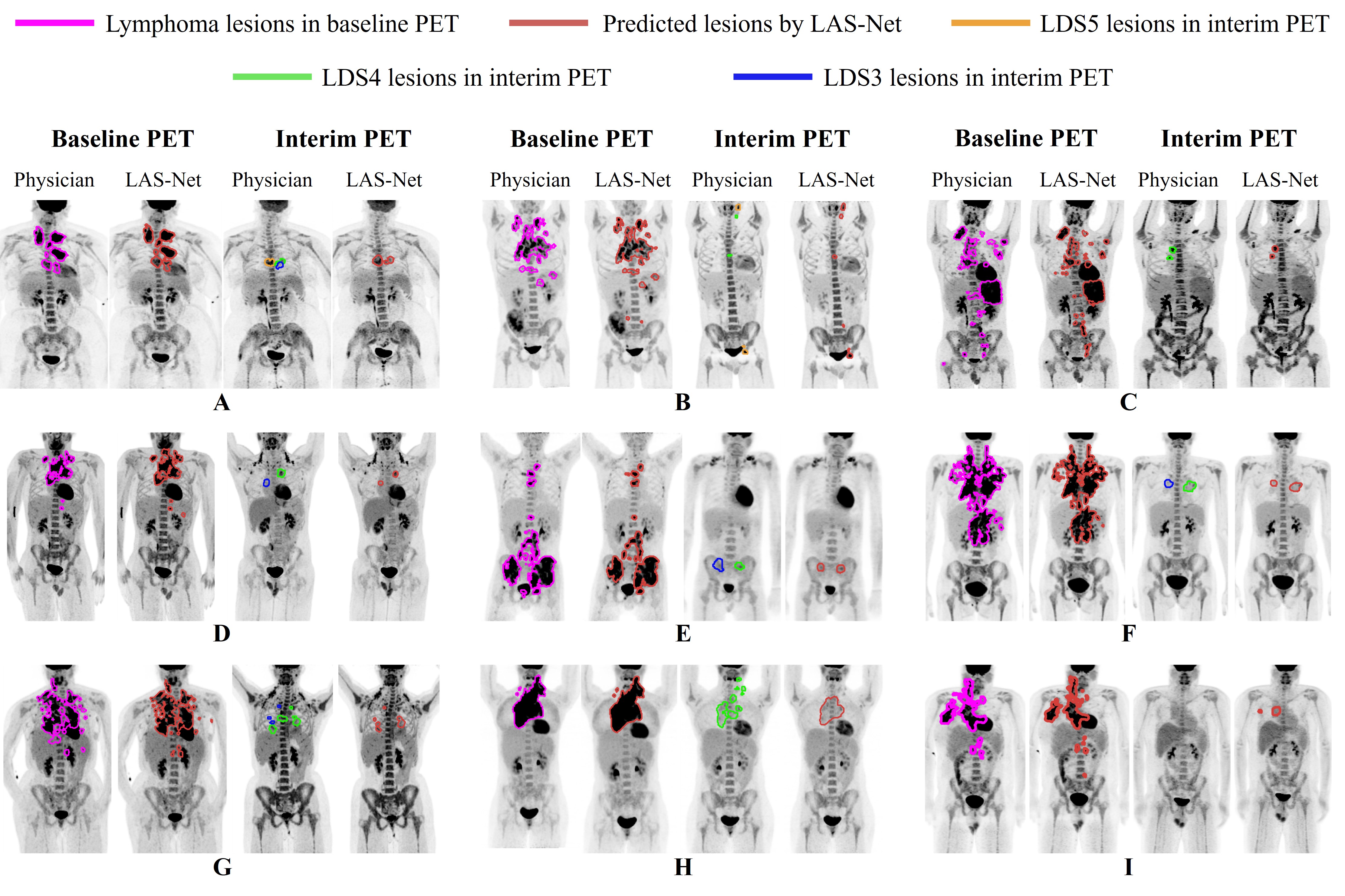}
  \caption{\small{Nine different examples of longitudinally-aware segmentation network (LAS-Net) output. Each case has maximum intensity projections (MIPs) of baseline and interim PET images with overlaying MIPs of the reference and predicted lesion masks. DS = Deauville score.}} 
    \vspace{-20pt}
  \label{fig:fig5}
\end{figure}

\subsection{Qualitative evaluation }
Figure \ref{fig:fig5} displays images from nine sample cases, each comprising baseline and interim lesion masks predicted by LAS-Net along with physician annotations. In cases A-F, LAS-Net successfully identified the residual lesions, including the lesions with the highest PET uptake (LDS4 or LDS5) as well as those with lower uptake (LDS3). Notably, in case B, LAS-Net detected new lesions, not present on PET1, located near the neck and bladder. If MPDR was applied, these true positive lesions would be excluded, leading to an underestimation of SUVmax and qPET.

In scenarios with multiple dispersed PET2 lesions (cases G-H), LAS-Net had difficulties in accurately identifying all lesions. Additionally, LAS-Net occasionally identified false positive lesions in negative cases (case I), especially when the residual SUVs were close to the mediastinum uptake. For baseline lymphoma segmentation, LAS-Net performed consistently well at delineating bulky diseases. Nonetheless, it was less effective in detecting small lesions situated at a distance from the primary disease sites, which was true for all comparator methods.

\begin{figure}[h!]
\vspace{-2pt}
\centering
\includegraphics[width=0.8\textwidth]{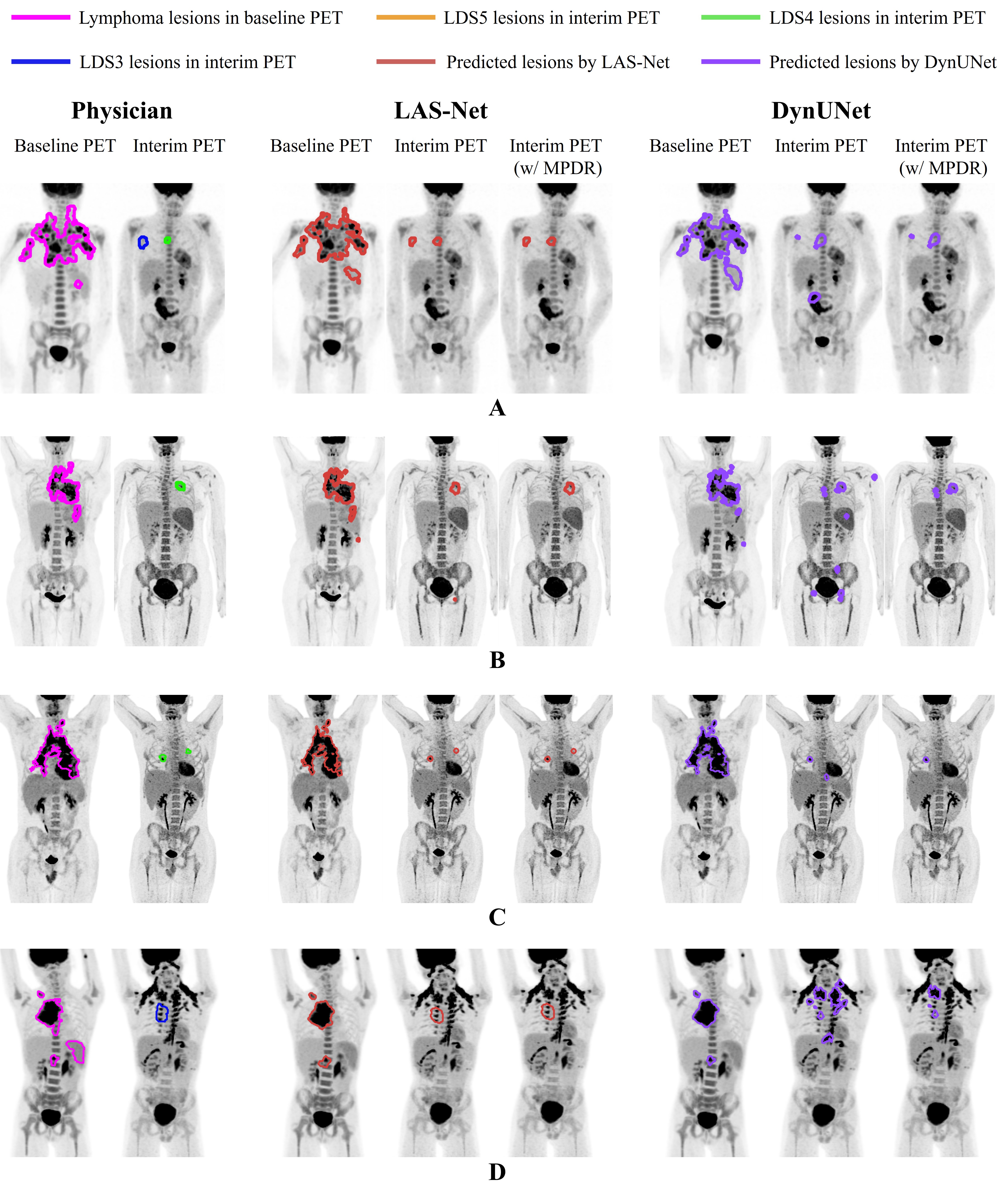}
  \caption{\small{Four examples comparing the proposed longitudinally-aware segmentation network (LAS-Net) with DynUNet, a model without longitudinal cross-attention. Each case has maximum intensity projections (MIPs) of baseline and interim PET images, overlaid with MIPs of reference and predicted lesion masks. For both LAS-Net and DynUNet output, results incorporating mask propagation through deformable registration (MPDR) are also included. DS = Deauville score.}} 
  \label{fig:fig6}
\end{figure}

To assess the benefits of integrating longitudinal awareness into the model architecture, we compared the predictions of LAS-Net with those of DynUNet in Figure \ref{fig:fig6}. Without applying MPDR, the PET2 false positives predicted by DynUNet significantly affected the accuracy of automated PET2 metrics. Especially in case D, DynUNet mistakenly identified brown fat uptake as residual lymphoma.

\subsection{Agreement of Model-Extract DS and Physician Assigned DS}
Table \ref{table:table3} presents patient-level DS classification results. If grouping cases into two categories – scores of DS 1, 2 vs. DS 3, 4, 5 – LAS-Net attained an F1 score of 0.752 (precision/recall: 0.687/0.836) and Cohen’s kappa of 0.630, outperforming (P<0.05) the top comparator, ST-Trans (with MPDR), which had 0.660 for F1 and 0.501 for Cohen’s kappa. If grouping based on DS of 1, 2 and 3 vs. DS 4 and 5, LAS-Net achieved an F1 score of 0.633 (precision/recall: 0.500/0.867) and Cohen’s kappa of 0.549, and was superior to other evaluated methods. 

\begin{table}[h!]
\vspace{-10pt}
\centering
\caption{Results of binary classification for adequate/inadequate treatment response using model-predicted Deauville scores.}
\includegraphics[width=0.8\textwidth]{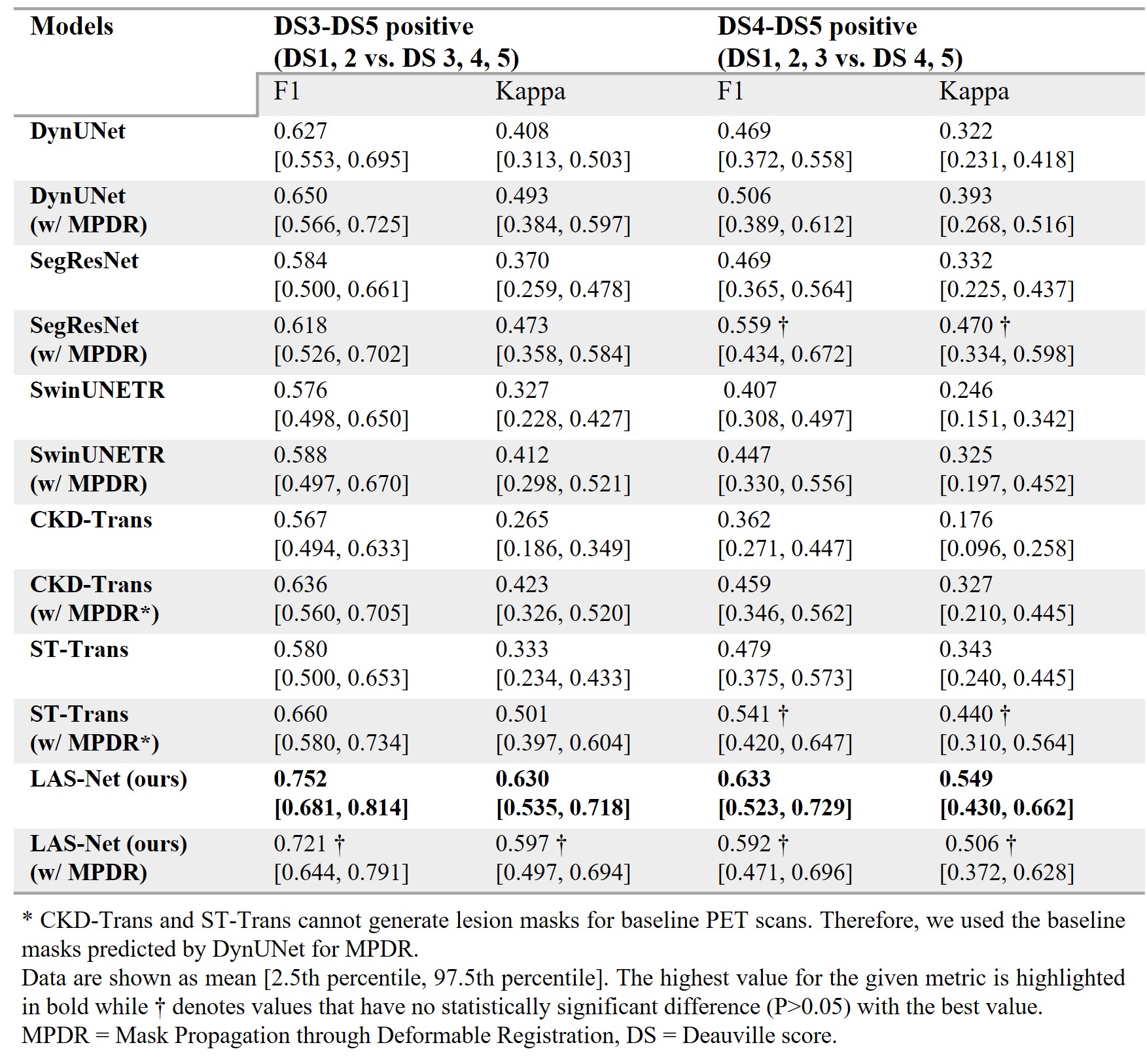}
\vspace{-10pt}
\label{table:table3}
\end{table}

\subsection{Ablation Studies }
The results of ablation studies are shown in Table \ref{table:table4}. We found that both LAWA and LAAG modules for longitudinal cross-attention improved lesion detection performance in PET2. Also, the inclusion of the PET1 branch and the combined PET1 and PET2 training enhanced the model’s capability to quantify PET2 scans. The choice of registration methods between PET1 and PET2 did not impact model performance. When input baseline and interim PET/CT images were co-registered using rigid registration, the performance was slightly worse but not significantly different from that achieved with deformable registration (P=0.22 for F1 scores).

\begin{table}[h!]
\vspace{10pt}
\centering
\caption{Ablation studies evaluating the effectiveness of each component in LAS-Net for interim lesion detection. }
\includegraphics[width=0.9\textwidth]{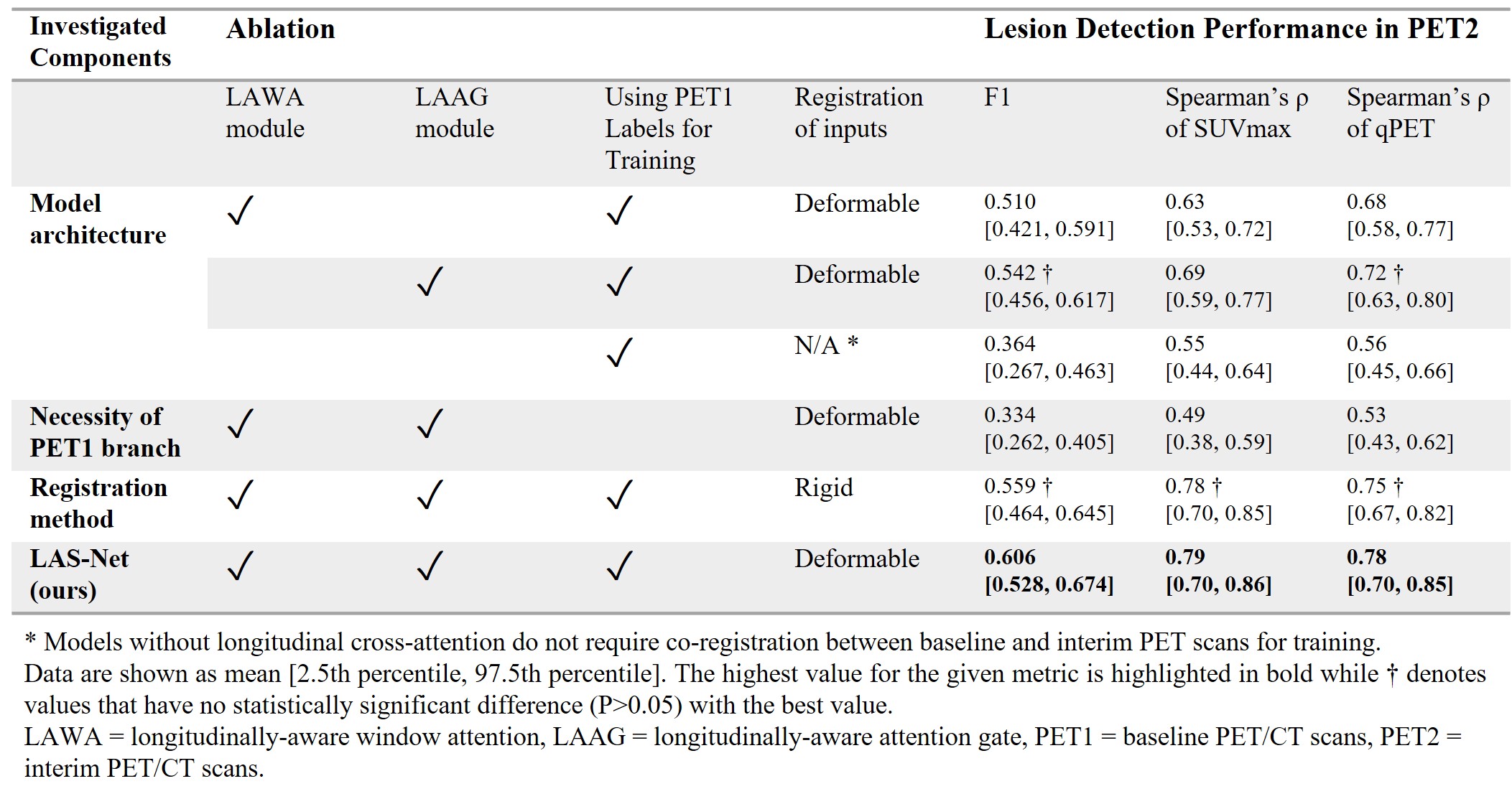}
\label{table:table4}
\end{table}

\subsection{External Testing}

We applied LAS-Net, trained on all AHOD1331 data, to the external AHOD0831 dataset. The detection F1 score in PET2 was 0.525 (95$\%$CI, 0.456, 0.582) and the Dice score in PET1 was 0.684 (95$\%$CI, 0.655, 0.711). Regarding quantitative PET metrics, the Spearman’s $\rho$ correlations between LAS-Net predictions and physician measurements showed a slight decrease: 0.70 for PET2 $\Delta$SUVmax, 0.69 for PET2 qPET, 0.87 for PET1 MTV and 0.89 for PET1 TLG. Detailed results along with example cases are provided in Appendix S6.

\newpage
\section{Discussion}
In this study, we introduced a novel deep-learning-based method (LAS-Net) for longitudinal analysis of serial PET/CT images in pediatric HL patients. Our approach was different from prior methods in two aspects. First, it used longitudinal cross-attention to extract baseline PET information for improved analysis of interim PET. Second, it adopted a dual-branch architecture to enable automatic quantification of both baseline and interim scans. Through comparative and ablation studies, we validated the effectiveness of our approach using data from two multi-center clinical trials, highlighting its potential to deliver rapid and consistent assessment of PET tumor burden and response.

Existing DL algorithms for detecting lymphoma lesions have been limited to analyzing PET1 scans without the ability to quantify PET2 for response assessment and outcome prediction. This limitation is primarily due to the challenge of detecting residual lymphoma in PET2, which often has low FDG uptake. It is even a difficult task for expert physicians, and they usually rely on PET1 (i.e., viewing PET1 and PET2 side-by-side) to identify residual lymphoma. Our method was intended to fill this gap by integrating longitudinal cross-attention mechanisms into the architecture. While previous research has leveraged prior PET data for interim image denoising (34) and response classification (35), our work distinguishes itself by incorporating longitudinal awareness to improve the analysis of multi-time-point imaging datasets. This design also aligns with the way physicians interpret PET scans over time. While our current study focuses on quantification, we believe that the principle underlying our method has broader applications. It can potentially be extended to the DL models aimed at diagnosis, especially for tasks requiring analysis of prior imaging data. It can also be adapted to accommodate multi-modal inputs by adding cross-attention modules in the intermediate layers. This may further improve the model’s accuracy but challenges, such as data alignment, increased computational complexity and the need for data fusion techniques would need to be addressed.

To develop a model for longitudinal response assessment in PET scans, we chose to jointly optimize our model for PET1 and PET2 analysis. This substantially improved model performance in identifying residual lesions in PET2, as evidenced by the ablation study. However, our model’s PET1 segmentation performance was no better than other SOTA models trained on PET1 scans. This was expected, as the PET1 branch should not, in principle, benefit from the PET2 branch.

The quantitative PET metrics that we investigated have been demonstrated to be better than visual criteria at guiding lymphoma treatment (6,7). For PET1, we found that MTV and TLG were the metrics most accurately quantified by the DL model. They are also the most time-consuming metrics for physicians to measure. Newly proposed distance-based metrics (Dmax, Dspleen) were harder for accurate quantification, because a single false positive or false negative can have a large impact on the values of these metrics. For PET2, we focused on measuring SUVmax, qPET, and the response metric $\Delta$SUVmax, as these have been associated with patient outcome in previous studies (23–25). Detecting residual lymphoma on PET2 was very challenging for models that did not use longitudinal information, and this was reflected in their poor F1 scores and their performance in PET2 quantification. Even with MPDR, these models were inferior to LAS-Net. 

This study has several limitations. First, despite showing clear advancements, we have not yet evaluated LAS-Net’s clinical utility, and there is still room for further improvements in its performance. Future work could explore advanced training strategies, such as pretraining or semi-supervised techniques, or use a larger labeled dataset. Second, we focused on quantitative PET metrics (MTV, qPET, etc.). Future research will aim to associate these metrics with patient outcome. Third, the labeling process for our external dataset differed from that used for our internal dataset. It is unclear if the performance drop in external testing is attributed to dataset shift, or different annotation quality. Additionally, this external cohort may not be large enough to evaluate the model’s real-world performance. Further investigation on a larger dataset collected from routine clinical practice is needed. Fourth, our current model only operates at two imaging time points. In future work, we hope to develop a unified framework that can process PET/CT images across all time points. Fifth, this study did not investigate how the predicted DSs impacted the Lugano classification, which could be an area of focus for future research. Lastly, we only evaluated our algorithm in the cohorts of pediatric HL patients. Whether it is applicable to other diseases or populations requires further investigation.

In conclusion, our study introduced a longitudinally-aware segmentation network to address the challenges of automatic quantification of serial PET scans. The proposed method demonstrated improved lesion detection performance in interim PET scans compared to other methods, without sacrificing segmentation accuracy in baseline PET scans. This highlights the critical role of incorporating longitudinal awareness into AI algorithms for tasks involving analyzing multi-time-point images.

\section*{Acknowledgments}
We sincerely thank Drs. Jihyun Kim and Inki Lee for annotating the AHOD0831 dataset in our previous study.

We acknowledge funding support from Imaging and Radiology Oncology Core Rhode Island (U24CA180803), Biomarker, Imaging and Quality of Life Studies Funding Program (BIQSFP), NIH (U10CA098543), NCTN Operations Center Grant (U10CA180886), NCTN Statistics $\&$ Data Center Grants (U10CA180899 and U10CA098413), QARC (CA29511), IROC RI (U24CA180803), and St. Baldrick's Foundation. 

Research reported in this publication was also supported by the National Institute Of Biomedical Imaging And Bioengineering of the National Institutes of Health under Award Number R01EB033782, by the Department of Defense under Award Number W81XWH-22-1-0336, and by GE Healthcare.

Disclaimer: The content is solely the responsibility of the authors and does not necessarily represent the official views of the National Institutes of Health.

\bibliographystyle{unsrt}  
\bibliography{references}  
\begin{itemize}[leftmargin=*]
\item[1.]  Mauz-Körholz C, Metzger ML, Kelly KM, et al. Pediatric Hodgkin Lymphoma. JCO. 2015;33(27):2975–2985. doi: \url{http://doi.org/10.1200/JCO.2014.59.4853}.
\item[2.]  Castellino SM, Pei Q, Parsons SK, et al. Brentuximab Vedotin with Chemotherapy in Pediatric High-Risk Hodgkin’s Lymphoma. N Engl J Med. 2022;387(18):1649–1660. doi: \url{http://doi.org/10.1056/NEJMoa2206660}.
\item[3.]  Friedman DL, Chen L, Wolden S, et al. Dose-Intensive Response-Based Chemotherapy and Radiation Therapy for Children and Adolescents With Newly Diagnosed Intermediate-Risk Hodgkin Lymphoma: A Report From the Children’s Oncology Group Study AHOD0031. JCO. 2014;32(32):3651–3658. doi: \url{http://doi.org/10.1200/JCO.2013.52.5410}. 
\item[4.] Kelly KM, Cole PD, Pei Q, et al. Response-adapted therapy for the treatment of children with newly diagnosed high risk Hodgkin lymphoma (AHOD0831): a report from the Children's Oncology Group. Br J Haematol. 2019;187(1):39-48. \url{http://doi.org/10.1111/bjh.16014}. 
\item[5.]  Cheson BD, Fisher RI, Barrington SF, et al. Recommendations for Initial Evaluation, Staging, and Response Assessment of Hodgkin and Non-Hodgkin Lymphoma: The Lugano Classification. JCO. 2014;32(27):3059–3067. doi: \url{http://doi.org/10.1200/JCO.2013.54.8800}.
\item[6.]  Rogasch JMM, Hundsdoerfer P, Hofheinz F, et al. Pretherapeutic FDG-PET total metabolic tumor volume predicts response to induction therapy in pediatric Hodgkin’s lymphoma. BMC Cancer. 2018;18(1):521. doi: \url{http://doi.org/10.1186/s12885-018-4432-4}. 
\item[7.]  Okuyucu K, Ozaydın S, Alagoz E, et al. Prognosis estimation under the light of metabolic tumor parameters on initial FDG-PET/CT in patients with primary extranodal lymphoma. Radiology and Oncology. 2016;50(4):360–369. doi: \url{http://doi.org/10.1515/raon-2016-0045}.
\item[8.]  Weisman AJ, Kieler MW, Perlman SB, et al. Convolutional Neural Networks for Automated PET/CT Detection of Diseased Lymph Node Burden in Patients with Lymphoma. Radiology: Artificial Intelligence. 2020;2(5):e200016. doi: \url{http://doi.org/10.1148/ryai.2020200016}. 
\item[9.]  Constantino CS, Leocádio S, Oliveira FPM, et al. Evaluation of Semiautomatic and Deep Learning–Based Fully Automatic Segmentation Methods on [18F]FDG PET/CT Images from Patients with Lymphoma: Influence on Tumor Characterization. J Digit Imaging. 2023;36(4):1864–1876. doi: \url{http://doi.org/10.1007/s10278-023-00823-y}.
\item[10.]  Huang L, Ruan S, Decazes P, Denœux T. Lymphoma segmentation from 3D PET-CT images using a deep evidential network. International Journal of Approximate Reasoning. 2022;149:39–60. doi: \url{http://doi.org/10.1016/j.ijar.2022.06.007}. 
\item[11.]  Hu H, Shen L, Zhou T, Decazes P, Vera P, Ruan S. Lymphoma Segmentation in PET Images Based on Multi-view and Conv3D Fusion Strategy. 2020 IEEE 17th International Symposium on Biomedical Imaging (ISBI). Iowa City, IA, USA: IEEE; 2020. p. 1197–1200. doi: \url{http://doi.org/10.1109/ISBI45749.2020.9098595}.
\item[12.]  Weisman AJ, Kim J, Lee I, et al. Automated quantification of baseline imaging PET metrics on FDG PET/CT images of pediatric Hodgkin lymphoma patients. EJNMMI Phys. 2020;7(1):76. doi: \url{http://doi.org/10.1186/s40658-020-00346-3}. 
\item[13.]  Yousefirizi F, Klyuzhin IS, O JH, et al. TMTV-Net: fully automated total metabolic tumor volume segmentation in lymphoma PET/CT images — a multi-center generalizability analysis. Eur J Nucl Med Mol Imaging. 2024; doi: \url{http://doi.org/10.1007/s00259-024-06616-x}.  
\item[14.]  Blanc-Durand P, Jégou S, Kanoun S, et al. Fully automatic segmentation of diffuse large B cell lymphoma lesions on 3D FDG-PET/CT for total metabolic tumour volume prediction using a convolutional neural network. Eur J Nucl Med Mol Imaging. 2021;48(5):1362–1370. doi: \url{http://doi.org/10.1007/s00259-020-05080-7}.
\item[15.] Weisman AJ, Lee I, Im H, et al. Machine learning-based assignment of Deauville scores is comparable to interobserver variability on interim FDG PET/CT images of pediatric lymphoma patients. Journal of Nuclear Medicine 61,1434–1434 (2020). 
\item[16.] Hatamizadeh A, Nath V, Tang Y, Yang D, Roth H, Xu D. Swin UNETR: Swin Transformers for Semantic Segmentation of Brain Tumors in MRI Images. arXiv; 2022. \url{http://arxiv.org/abs/2201.01266}. Accessed February 8, 2023
\item[17.]  Liu Z, Lin Y, Cao Y, et al. Swin Transformer: Hierarchical Vision Transformer using Shifted Windows. 2021 IEEE/CVF International Conference on Computer Vision (ICCV). Montreal, QC, Canada: IEEE; 2021. p. 9992–10002. doi: \url{http://doi.org/10.1109/ICCV48922.2021.00986}. 
\item[18.] Oktay O, Schlemper J, Folgoc LL, et al. Attention U-Net: Learning Where to Look for the Pancreas. arXiv; 2018. doi: \url{http://doi.org/10.48550/arXiv.1804.03999}. Accessed February 8, 2023
\item[19.]  Im H-J, Bradshaw T, Solaiyappan M, Cho SY. Current Methods to Define Metabolic Tumor Volume in Positron Emission Tomography: Which One is Better? Nucl Med Mol Imaging. 2018;52(1):5–15. doi: \url{http://doi.org/10.1007/s13139-017-0493-6}. 
\item[20.]  Larson SM, Erdi Y, Akhurst T, et al. Tumor Treatment Response Based on Visual and Quantitative Changes in Global Tumor Glycolysis Using PET-FDG Imaging. The Visual Response Score and the Change in Total Lesion Glycolysis. Clin Positron Imaging. 1999;2(3):159-171. doi: \url{http://doi.org/10.1016/s1095-0397(99)00016-3}. 
\item[21.]  Cottereau A-S, Nioche C, Dirand A-S, et al. 18 F-FDG PET Dissemination Features in Diffuse Large B-Cell Lymphoma Are Predictive of Outcome. J Nucl Med. 2020;61(1):40–45. doi: \url{http://doi.org/10.2967/jnumed.119.229450}.
\item[22.]  Girum KB, Cottereau A-S, Vercellino L, et al. Tumor Location Relative to the Spleen Is a Prognostic Factor in Lymphoma Patients: A Demonstration from the REMARC Trial. J Nucl Med. 2024;65(2):313–319. doi: \url{http://doi.org/10.2967/jnumed.123.266322}. 
\item[23.]  Hasenclever D, Kurch L, Mauz-Körholz C, et al. qPET – a quantitative extension of the Deauville scale to assess response in interim FDG-PET scans in lymphoma. Eur J Nucl Med Mol Imaging. 2014;41(7):1301–1308. doi: \url{http://doi.org/10.1007/s00259-014-2715-9}.
\item[24.]  Santos FM, Marin JFG, Lima MS, et al. Impact of baseline and interim quantitative PET parameters on outcomes of classical Hodgkin Lymphoma. Ann Hematol. 2023; doi: \url{http://doi.org/10.1007/s00277-023-05461-6}.
\item[25.]  Yang S, Qiu L, Huang X, Wang Q, Lu J. The prognostic significance of $\Delta$SUVmax assessed by PET/CT scan after 2 cycles of chemotherapy in patients with classic Hodgkin’s lymphoma. Ann Hematol. 2020;99(2):293–299. doi: \url{http://doi.org/10.1007/s00277-019-03892-8}.
\item[26.]  Isensee F, Jaeger PF, Kohl SAA, Petersen J, Maier-Hein KH. nnU-Net: a self-configuring method for deep learning-based biomedical image segmentation. Nat Methods. 2021;18(2):203–211. doi: \url{http://doi.org/10.1038/s41592-020-01008-z}.
\item[27.] Cardoso MJ, Li W, Brown R, et al. MONAI: An open-source framework for deep learning in healthcare. arXiv; 2022. \url{http://arxiv.org/abs/2211.02701}. Accessed March 1, 2023.
\item[28.]  Myronenko A. 3D MRI brain tumor segmentation using autoencoder regularization. arXiv; 2018. \url{http://arxiv.org/abs/1810.11654}. Accessed February 19, 2024
\item[29.] Lin J, Lin J, Lu C, et al. CKD-TransBTS: Clinical Knowledge-Driven Hybrid Transformer With Modality-Correlated Cross-Attention for Brain Tumor Segmentation. IEEE Trans Med Imaging. 2023;42(8):2451–2461. doi: \url{http://doi.org/10.1109/TMI.2023.3250474}.
\item[30.]  Zhang J, Cui Z, Shi Z, et al. A robust and efficient AI assistant for breast tumor segmentation from DCE-MRI via a spatial-temporal framework. Patterns. 2023;4(9):100826. doi: \url{http://doi.org/10.1016/j.patter.2023.100826}.
\item[31.] 	Huff DT, Santoro-Fernandes V, Chen S, et al. Performance of an automated registration-based method for longitudinal lesion matching and comparison to inter-reader variability. Phys Med Biol. 2023;68(17):175031. doi: \url{http://doi.org/10.1088/1361-6560/acef8f}.
\item[32.]  Barrington SF, Kluge R. FDG PET for therapy monitoring in Hodgkin and non-Hodgkin lymphomas. Eur J Nucl Med Mol Imaging. 2017;44(S1):97–110. doi: \url{http://doi.org/10.1007/s00259-017-3690-8}.
\item[33.]  Smith L, Tanabe LK, Ando RJ nee, et al. Overview of BioCreative II gene mention recognition. Genome Biol. 2008;9(S2):S2. doi: \url{http://doi.org/10.1186/gb-2008-9-s2-s2}.
\item[34.] Wang Y-R (Joyce), Qu L, Sheybani ND, et al. AI Transformers for Radiation Dose Reduction in Serial Whole-Body PET Scans. Radiology: Artificial Intelligence. 2023;5(3):e220246. doi: \url{http://doi.org/10.1148/ryai.220246}.
\item[35.] Joshi A, Eyuboglu S, Huang S-C, et al. OncoNet: Weakly Supervised Siamese Network to automate cancer treatment response assessment between longitudinal FDG PET/CT examinations. arXiv; 2021. \url{http://arxiv.org/abs/2108.02016}. Accessed January 9, 2024.
\end{itemize}

\clearpage
\titleformat*{\section}{\normalfont\large\bfseries\centering}
\section*{Supplementary Material}
\titleformat*{\section}{\normalfont\large\bfseries} 

\subsection*{Appendix S1. Labeling Procedures}
Details of our labeling approach can be found in the labeling guide available at \url{https://github.com/xtie97/lymphoma_labeling_guide}. In short, lymphoma detection in the internal AHOD1331 cohort was facilitated by a customized MIM LesionID workflow. A standardized uptake value (SUV) and volume (2 ml) threshold was first used to pre-identify regions of high FDG uptake based on the PERCIST criteria (1). Then the annotator deleted regions of interest (ROIs) that did not contain tumors. Any tumor regions that were missed by pre-labelling (such as lesions < 2 ml) were manually added by the annotator using the PET Edge+ tool in the MIM software. For any liver and osseous/bone marrow involvement, only focal diseases were identified. Splenic lesions were considered if focal uptake was present or diffuse uptake was higher than 1.5 of the liver SUV. When it was unclear whether a lesion was lymphoma or physiological, it was classified as “equivocal”. After the first annotator labeled all cases, the second annotator (one of two senior nuclear medicine physicians) reviewed and edited the contours as necessary. Given the absence of a universally-accepted approach for delineation of tumor boundaries (2), we performed an internal calibration study to evaluate various PET thresholding methods against a set of physician-drawn contours. Consequently, we used a union of SUV $>$ 2.5 and SUV $>$ 40$\%$ of SUVmax within the lesion ROIs to create final segmentation masks. To analyze interim PET scans, the annotator compared baseline and interim PET side-by-side and used PET Edge+ to add residual tumors.  

In the external AHOD0831 cohort, the annotator placed large ROIs around areas containing disease using Mirada XD software, excluding diffused osseous/bone marrow involvement and regions with physiological uptake. Then a union of SUV $\>$ 2.5 and SUV $\>$ 40$\%$ of SUVmax was applied within each lesion ROI for segmentation of the lymphoma disease. For interim PET scans, the annotator manually added the residual lesions that had FDG activity above mediastinum uptake. 

\subsection*{Appendix S2. Image Preprocessing}
In both internal and external cohorts, we resampled PET and CT images to a voxel size of 3$\times$3$\times$3 mm using trilinear interpolation. Labels were resampled to the same voxel size via nearest neighbor interpolation. To reduce the spatial discrepancies across longitudinal imaging data, we registered the baseline scans to the interim CT using deformable transformation. We used ANTsPy (0.4.2), a python library for medical image registration. Considering that there is a lot of background information near the edge, we cropped the PET and CT volumes using bounding boxes determined by a SUV threshold of 0.2. PET SUVs and CT Hounsfield units (HUs) were then linearly scaled from [0,30] to [0,1] and from [-150, 250] to [0,1], respectively. During training, equivocal and non-equivocal lesions were combined and used as the ground truth mask. 

\subsection*{Appendix S3. Model Training and Inference}
We first concatenated the PET and CT images as two channels for model input and then cropped random patches of 112$\times$112$\times$112 centered on the areas of lesion class with a probability of 8/9 (1/9 for background). To alleviate the over-fitting problem, we applied the data augmentation techniques to training data, including random affine transformation (rotation between -25 and 25 degrees, axis flip for all dimensions, zoom between 0.8 to 1.2), Gaussian noise and Gaussian blur. We jointly optimized the baseline and interim PET branches, using the following loss function:
\begin{align*}
    L(y_1, y_2, x_1, x_2; g_1,g_2) = & \Bigl(L_{CE} \bigl(y_1, g_1(x_1)\bigr) + L_{Dice} \bigl(y_1, g_1(x_1)\bigr)\Bigr) + \\ & \Bigl(L_{CE} \bigl(y_2, g_2(x_1, x_2)\bigr) + L_{Dice} \Bigl(y_2, g_2\big(x_1,x_2)\bigr)\Bigr) 
\end{align*}
    
Where $x_1$, $x_2$ denote baseline PET/CT (PET1) and interim PET/CT (PET2). $y_1$, $y_2$ denote reference baseline and interim lesion masks. $g_1$, $g_2$ denote the PET1 and PET2 branches of the model. Note that $g_1$ solely depends on PET1 while $g_2$ takes inputs from both PET1 and PET2. For each branch, the loss is an unweighted sum of cross-entropy (CE) loss and Dice loss, which has proven effective in various segmentation tasks (3). To enable the model to learn joint feature representations from both time points, all components and weights in the model are shared between the two branches, except for the longitudinal cross-attention (LCA). Specifically, weight sharing is applied at each level between the PET1 and the PET2 branches. This includes convolutional blocks, multi-head self-attention and self-attention gate blocks. The LCA mechanism is designed to integrate features from PET1 into the analysis of PET2, but not vice versa. In this setup, the output features from the self-attention blocks in both branches (denoted as $z_1^{l-1}$ for PET1 and $z_2^{l-1}$ for PET2) are fed into the LCA mechanism (denoted as $f_{LCA}$), with its output being added to the PET2 features. 
\begin{align*}
    z_2^l = z_2^{l-1} + f_{LCA} (z_1^{l-1}, z_2^{l-1})
\end{align*}

Where $z_2^l$ denotes the updated PET2 features. For the PET1 branch, the longitudinal cross-attention does not modify its features, preventing any reverse information flow from PET2 to PET1. Therefore, the PET1 features remain unchanged: 
\begin{align*}
    z_1^l = z_1^{l-1}
\end{align*}

The models were trained using the AdamW optimizer (4), with an initial learning rate of $10^{-4}$, weight decay regularization of $10^{-5}$, and a cosine annealing scheduler. We set the batch size to 8 and trained the models for 300 epochs on NVIDIA A100 GPUs. The learning environment requires the following Python (3.8.8) libraries: PyTorch (1.13.0), Monai (1.3.0).

During inference, we generated lesion masks for baseline and interim PET scans separately. For baseline mask prediction, original PET1 scans were used as input for the PET1 branch, while PET2 branch inputs can be set to any value (e.g., zeros or random noise) since the longitudinal cross-attention in LAS-Net is unidirectional. However, interim mask predictions require both PET1 and PET2 scans as input, with PET1 being deformable-registered to PET2. We employed the sliding window method with an overlap rate of 0.625 and blended outputs of overlapping patches using Gaussian weighting. To generate the binary segmentation mask, we applied a threshold of 0.5 to the model’s output followed by removing any small region with a volume below 0.2 ml using connected component analysis.

\subsection*{Appendix S4. Quantitative PET metrics}
\begin{table}[h!]
\vspace{-2pt}
\centering
\caption*{Table E1: Definitions of quantitative PET metrics included in this study.}
\includegraphics[width=0.8\textwidth]{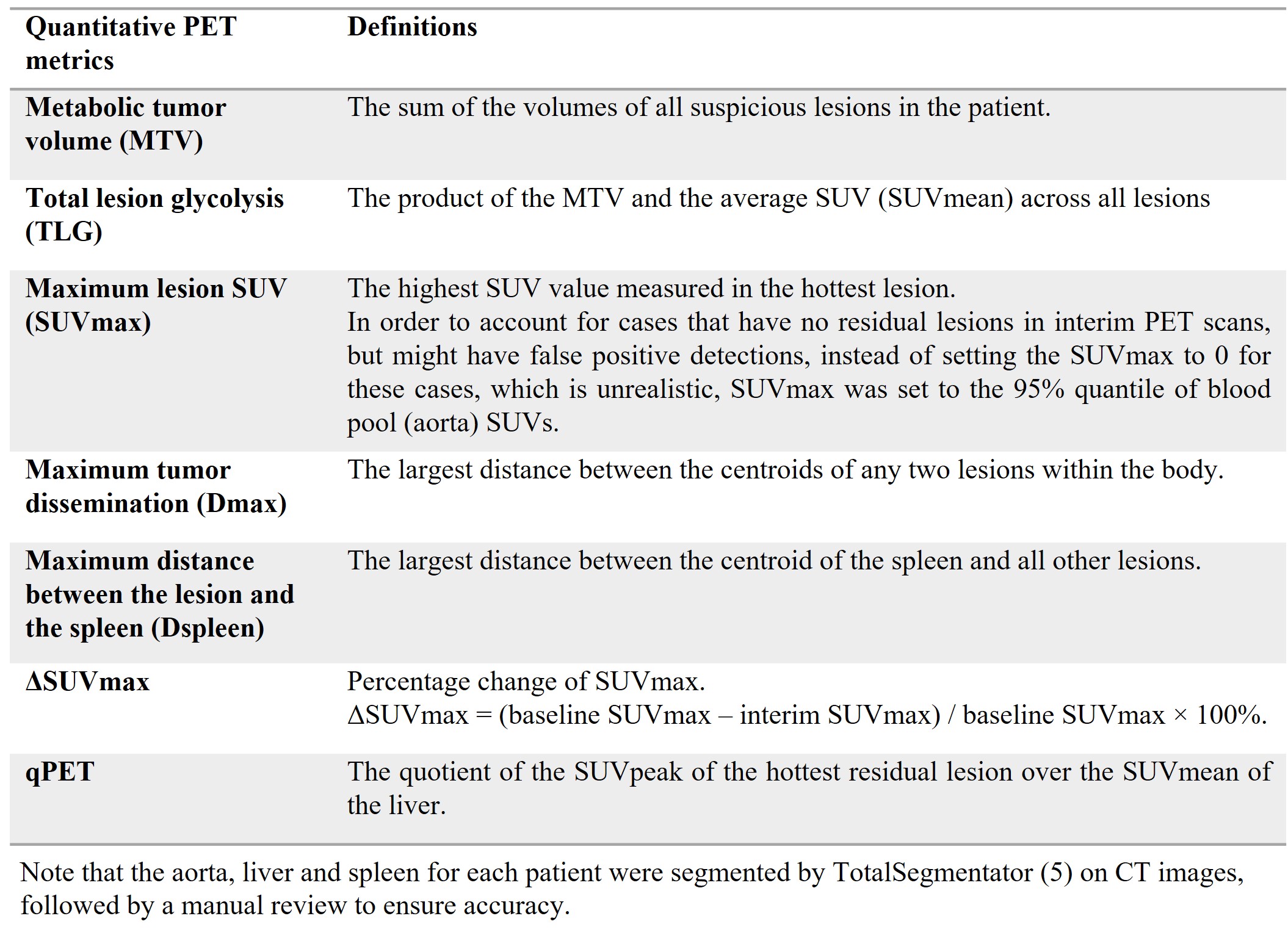}
\label{table:TableE1}
\end{table}

\subsection*{Appendix S5. Results with Equivocal Lesions Included} 
Figure E1 presents the comparison results for lesion detection on interim PET scans when equivocal lesions are included. The F1 scores do not exhibit statistically significant differences with those computed for non-equivocal lesions only.

\begin{figure}[t!]
\centering
\includegraphics[width=0.98\textwidth]{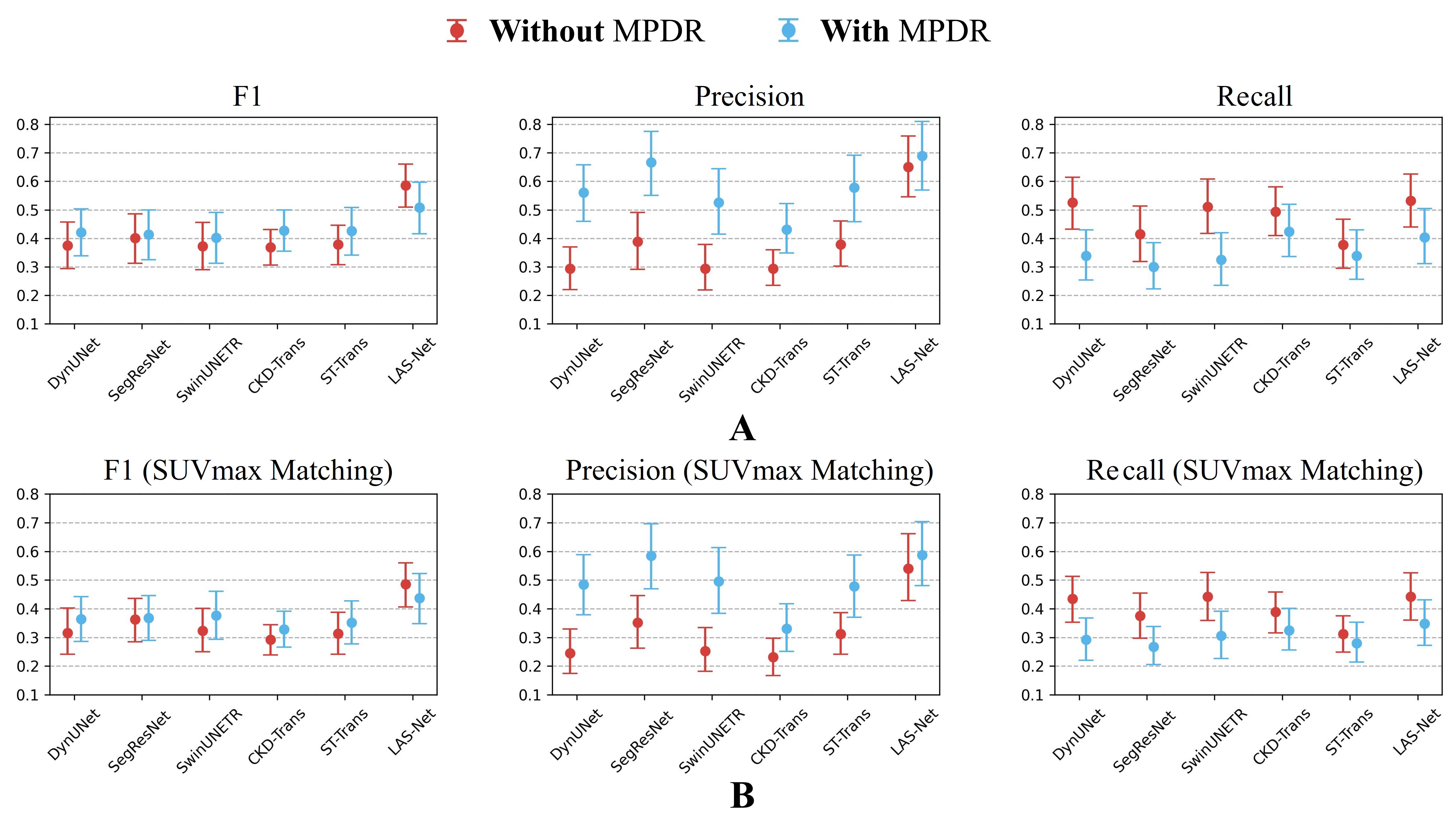}
  \caption*{Figure E1: \small{Performance comparison of interim PET lesion detection in the internal cohort with the inclusion of equivocal lesions. Results are reported with and without mask propagation through deformable registration (MPDR). Both (A) and (B) show the results of detection F1 scores, precision, and recall but they adopt different criteria for classifying true positives. (A) uses the criterion that a predicted lesion is considered as a true positive if it overlaps with at least one voxel of the reference lesion. (B) uses the criterion that the predicted lesion’s SUVmax should be matched with the reference lesion’s SUVmax for it to be considered a true positive. In the plots, actual metric values are marked by circles with error bars indicating 95$\%$ confidence intervals.}} 
  \label{fig:figE1}
\end{figure}

\subsection*{Appendix S6. External Testing}
Figure E2 shows the scatter plots comparing quantitative PET metrics measured by our model and by physicians in the external AHOD0831 dataset. Figure E3 presents six sample cases from the AHOD0831 dataset, each comprising model predictions and reference physician annotations.

\begin{figure}[t!]
\centering
\includegraphics[width=0.79\textwidth]{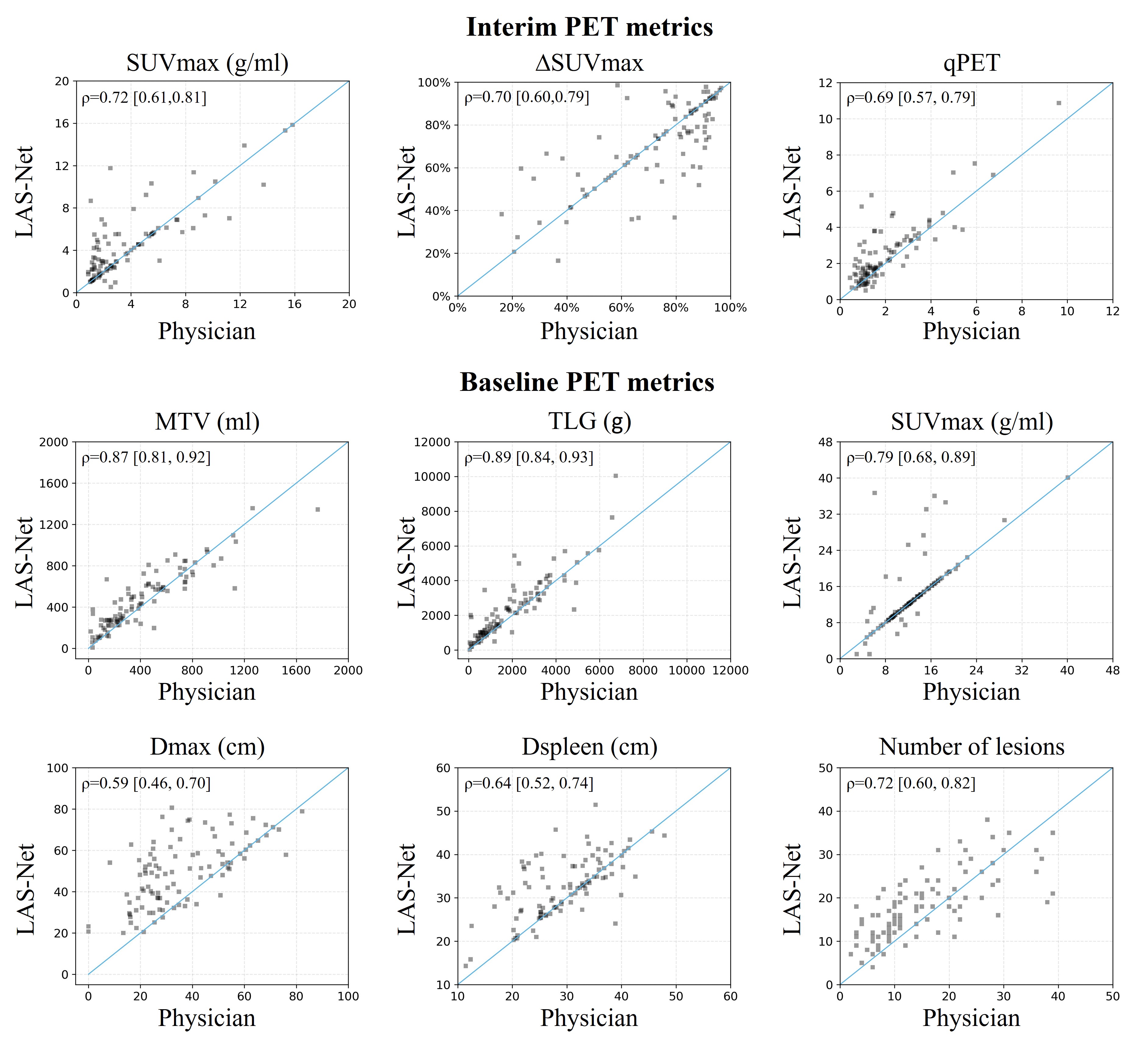}
  \caption*{Figure E2: \small{Comparison of physician-based and automatically extracted PET metrics in the external AHOD0831 cohort. Spearman’s $\rho$ correlations are shown in the top left corner of each plot. Correlation values are presented as mean [2.5th percentile, 97.5th percentile].}} 
  \label{fig:figE2}
\end{figure}

\begin{figure}[h!]
\centering
\includegraphics[width=0.95\textwidth]{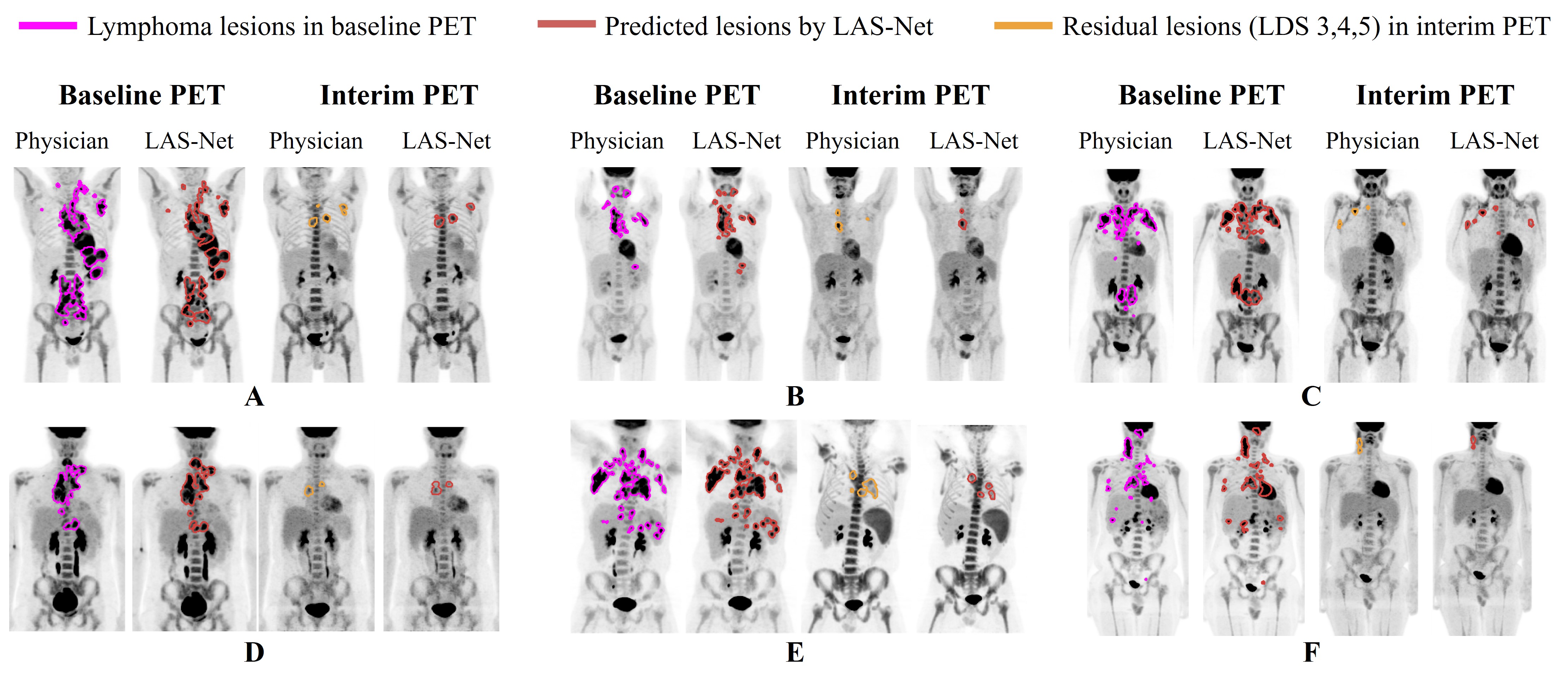}
  \caption*{Figure E3: \small{Six examples of longitudinally-aware segmentation network (LAS-Net) output in the external AHOD0831 cohort. Each case has maximum intensity projections (MIPs) of baseline and interim PET images with overlaying MIPs of the reference and predicted lesion masks. Note that lesion-level Deauville scores are not available for the AHOD0831 data. DS = Deauville score}} 
  \label{fig:figE3}
\end{figure}

\subsection*{Appendix S7. Performance of Interim Lesion Detection Using the Criterion of Dice$>$0.5}

To provide a more comprehensive analysis of the model’s detection performance on interim PET scans, we defined an even more stringent criterion: a predicted lesion was considered as a true positive when this lesion had a Dice coefficient above 0.5 with a true lesion, otherwise the predicted lesion was classified as a false positive and the true lesion was a false negative.

With this criterion (results presented in Figure E4), LAS-Net attained an F1 score of 0.412 (95$\%$CI, 0.337, 0.490). All comparator methods' F1 scores were also decreased, with ST-Trans (with MPDR) achieving the highest value (0.319, 95$\%$CI, 0.240, 0.403).

\begin{figure}[h!]
\centering
\includegraphics[width=0.98\textwidth]{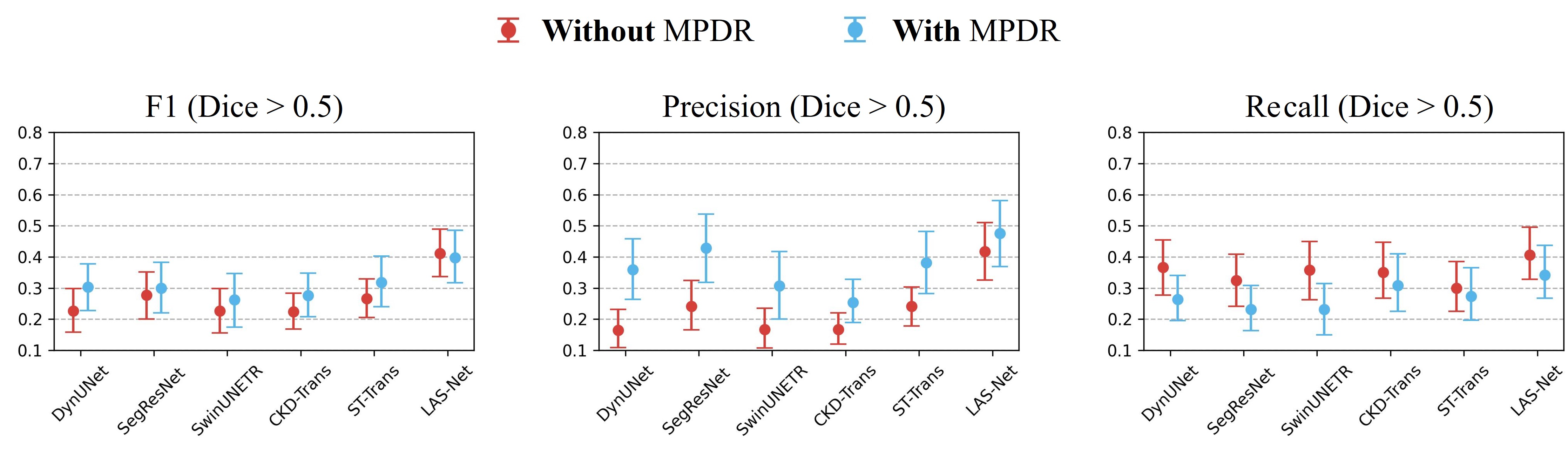}
  \caption*{Figure E4: \small{Performance comparison of interim PET lesion detection in the internal cohort using the criterion that a predicted lesion is classified as a true positive if the Dice coefficient between the predicted lesion and the reference lesion exceeds 0.5.}} 
  \label{fig:figE4}
\end{figure}

\subsection*{Appendix S8. Subgroup Analysis}
Figure E5 shows the results of subgroup analyses based on age, sex, patient weight, normalized injected dose (i.e., injected dose divided by patient weight) and scanner models (including two most representative manufactures in the internal and external cohorts, i.e., GE Healthcare and Siemens). We also evaluated model performance on scans from overlapping scanner models (i.e., present in both internal and external cohorts) versus non-overlapping models (i.e., present in only one cohort).

First, we assessed performance differences between subgroups within each cohort. In the internal cohort, no significant differences were observed for interim PET across any subgroup. For baseline PET, the Dice score was significantly higher in the group older than 15 years compared to those 15 years and younger (P=0.045). No significant differences were found for other characteristics. In the external cohort, no significant differences were noted between subgroups for any characteristic.

We then compared the performance between the internal and external cohorts for each subgroup. For interim PET, statistically significant performance differences between the internal and external cohorts were found in the following subgroups: age $\leq$ 15 years, female, weight $\leq$ 60 kg, normalized injected dose $>$ 5.5 MBq/kg and Siemens. For baseline PET, the internal performance was consistently higher than the external performance across all subgroups by a significant margin.

\begin{figure}[h!]
\centering
\includegraphics[width=0.9\textwidth]{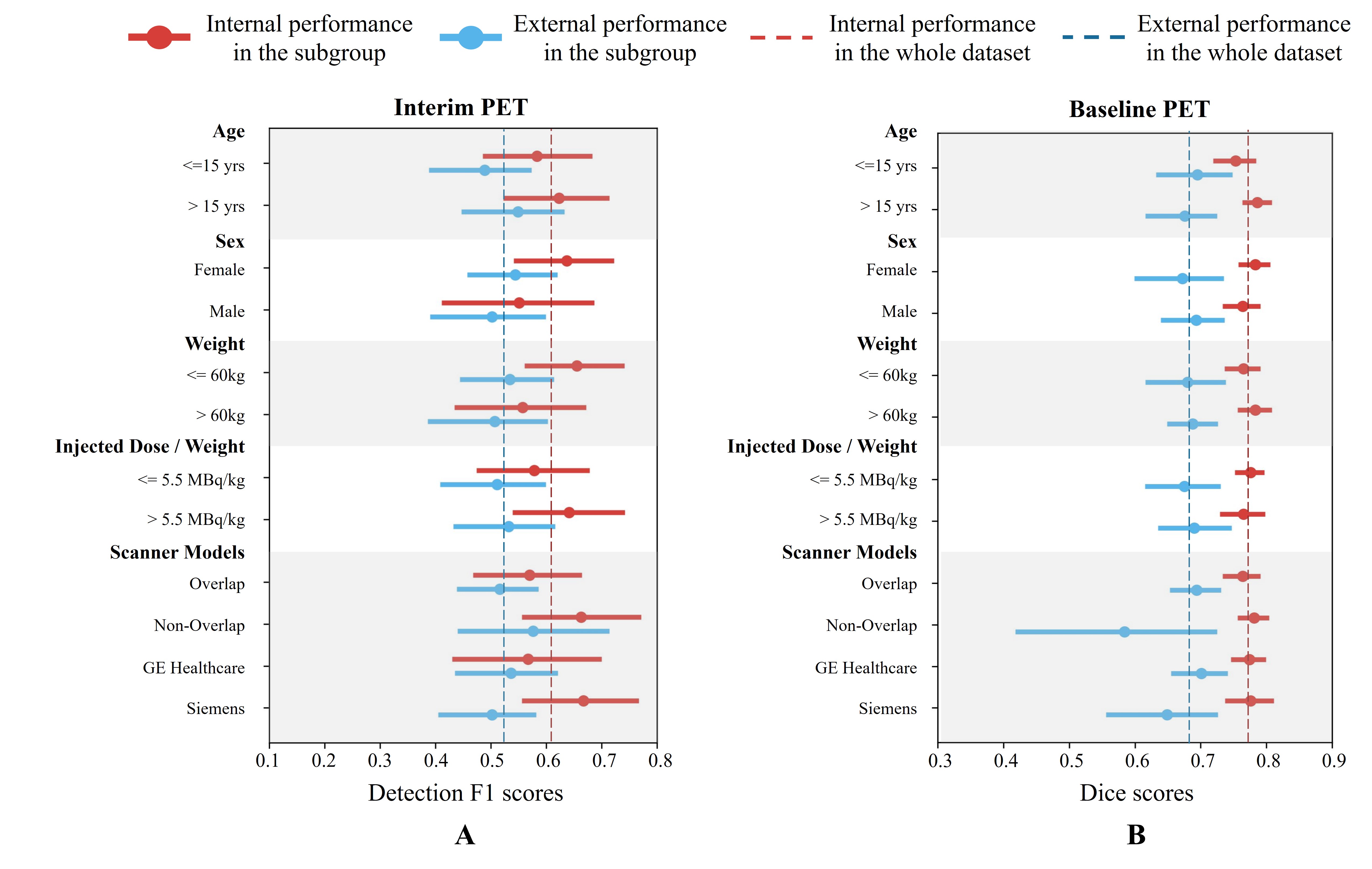}
  \caption*{Figure E5: \small{Internal and external performance across subgroups based on age, sex, weight, normalized injected dose, and scanner models. (A) shows lesion detection performance on interim PET across subgroups, and (B) shows lesion segmentation performance on baseline PET across subgroups. Each metric value is marked by a circle, with error bars indicating the 95$\%$ confidence intervals. It is important to note that for the external cohort, there were only 6 scans from non-overlapping scanner models (i.e., these models were only present in the external cohort), which resulted in large error bars.}} 
  \label{fig:figE5}
\end{figure}

\clearpage
\subsection*{Appendix S9. Comparison between the internal cohort and the full AHOD1331 study cohort}

\begin{table}[h!]
\vspace{0pt}
\centering
\caption*{Table E2: Demographics and scan characteristics of our internal and the full AHOD1331 study cohorts.}
\includegraphics[width=0.9\textwidth]{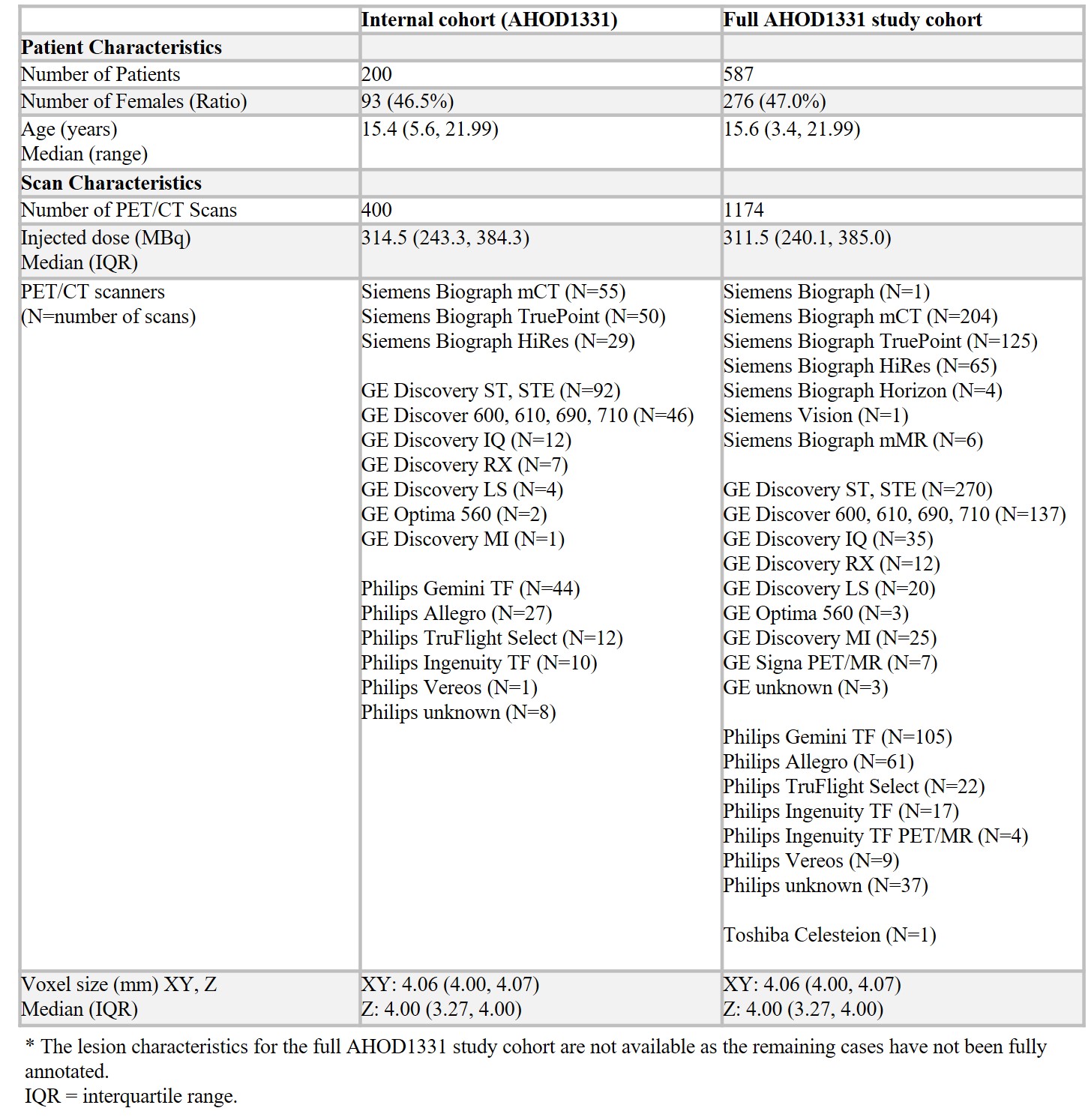}
\label{table:TableE2}
\end{table}

\clearpage
\section*{References}
\begin{itemize}[leftmargin=*]
\item [1.] Wahl RL, Jacene H, Kasamon Y, Lodge MA. From RECIST to PERCIST: Evolving Considerations for PET response criteria in solid tumors. J Nucl Med. 2009;50 Suppl 1(Suppl 1):122S-50S. doi: \url{http://doi.org/10.2967/jnumed.108.057307}.
\item [2.]	Martín-Saladich Q, Reynés-Llompart G, Sabaté-Llobera A, Palomar-Muñoz A, Domingo-Domènech E, Cortés-Romera M. Comparison of different automatic methods for the delineation of the total metabolic tumor volume in I-II stage Hodgkin Lymphoma. Sci Rep. 2020;10(1):12590. doi: \url{http://doi.org/10.1038/s41598-020-69577-9}.
\item [3.] Ma J, He Y, Li F, Han L, You C, Wang B. Segment anything in medical images. Nat Commun. 2024;15(1):654. doi: \url{http://doi.org/10.1038/s41467-024-44824-z}.
\item [4.]	Loshchilov I, Hutter F. Decoupled Weight Decay Regularization. arXiv; 2019. \url{http://arxiv.org/abs/1711.05101}. Accessed August 31, 2023.
\item [5.] Wasserthal J, Breit H-C, Meyer MT, et al. TotalSegmentator: Robust Segmentation of 104 Anatomic Structures in CT Images. Radiology: Artificial Intelligence. Radiological Society of North America; 2023;5(5):e230024. doi: \url{http://doi.org/10.1148/ryai.230024}.

\item [6.] Ahamed S, et al. Comprehensive Evaluation and Insights into the Use of Deep Neural Networks to Detect and Quantify Lymphoma Lesions in PET/CT Images. arXiv; 2023. \url{http://arxiv.org/abs/2311.09614}. Accessed December 6, 2023.

\end{itemize}

\end{document}